\documentclass[conference]{IEEEtran}
\IEEEoverridecommandlockouts

\usepackage{cite}
\usepackage{amsmath,amssymb,amsfonts}
\usepackage{algorithmic}
\usepackage{graphicx}
\usepackage{textcomp}
\usepackage{xcolor}
\usepackage{comment}

\usepackage{booktabs}
\usepackage{array}
\usepackage{multirow}
\usepackage{enumitem}
\usepackage{float}
\usepackage{epstopdf}      
\DeclareRobustCommand{\ScheduleFreePlus}{ScheduleFree\raisebox{0.2ex}{\textnormal{\footnotesize +}}}

\usepackage{flushend}

\usepackage{fancyhdr}

\def\BibTeX{{\rm B\kern-.05em{\sc i\kern-.025em b}\kern-.08em
    T\kern-.1667em\lower.7ex\hbox{E}\kern-.125emX}}

\begin{document}

\title{Development of Deep Learning Optimizers: Approaches, Concepts, and Update Rules}

\author{
    \IEEEauthorblockN{Dogay Altinel}
    \IEEEauthorblockA{\textit{Department of Electrical and Electronics Engineering} \\ \textit{Istanbul Medeniyet University} \\
    Istanbul, Turkiye \\
    dogay.altinel@medeniyet.edu.tr}
}

\maketitle

\thispagestyle{fancy}

\begin{abstract}
Deep learning optimizers are optimization algorithms that enable deep neural networks to learn. The effectiveness of learning is highly dependent on the optimizer employed in the training process. Alongside the rapid advancement of deep learning, a wide range of optimizers with different approaches have been developed. This study aims to provide a review of various optimizers that have been proposed and received attention in the literature. The optimizers are examined individually in chronological order, from Stochastic Gradient Descent and Momentum to recent methods such as AdamW, Sophia, Muon, and \ScheduleFreePlus, and their distinctive features are highlighted. The update rule of each optimizer is presented in detail, with an explanation of the associated concepts and variables. The techniques applied by these optimizers, their contributions to the optimization process, and their default hyperparameter settings are also discussed. To complement the review, visual trajectory analyses are included as illustrative examples of optimizer behavior on a controlled benchmark landscape. A small-scale transformer language-modeling experiment is also included to provide a limited task-level complement to the mechanism-oriented analysis. In addition, insights are offered into the open challenges encountered in the optimization of deep learning models. Thus, a comprehensive resource is provided both for understanding the current state of optimizers and for identifying potential areas of future development.
\end{abstract}


\section{Introduction}
In recent years, artificial intelligence (AI) has gained significant importance, with deep learning models emerging as a central component of the field \cite{goodfellow2016}. Deep learning models based on artificial neural networks are trained to perform specific tasks. The training process of these models must be optimized with respect to convergence stability, generalization performance, and computational cost \cite{chen2018,charles2018,sun2019a}. To this end, deep learning optimization algorithms, also known as optimizers, are employed during the training process to ensure that the neural networks underlying the models are able to learn effectively from data \cite{sun2019b,shrestha2019}. Deep learning optimizers are optimization algorithms used for neural networks to minimize the loss function iteratively. In this way, the goal is to indirectly improve model performance metrics that are difficult to optimize directly. The loss function in deep learning corresponds to the objective function in any optimization problem. Depending on the task that the model performs, various functions such as Mean squared error, Mean absolute error, Binary cross entropy, Categorical cross entropy, etc.\ can be selected as the loss function \cite{terven2023}. In deep learning, the forward pass is the processing of the input data samples on the neural network from the first layer to the last. At the end of the forward pass, the loss function is used to compute the distance between the deep learning model's predicted values and the actual target values. The values resulting from computations performed on each data sample during a training epoch are commonly averaged. In this sense, a loss value is obtained that measures how well the model works.

The optimizers determine how to change the loss function parameters to reduce the loss value during the training of deep learning models, which directly affects both the training process and model performance \cite{zaheer2019,hassan2023,altinel2025performance}. Here, loss function parameters are the deep learning model parameters that can be changed by the optimizer, i.e., trainable parameters. Weights, kernel weights, embedding weights, iterative weights, attention weights, batch normalization parameters, activation function parameters, and biases are commonly encountered trainable parameters in deep learning models. These parameters, which can number in the millions, are updated step by step with optimizers using the backpropagation method. Backpropagation is the process of updating parameters by typically using gradient computations from the output layer to the previous layers \cite{werbos1974,rumelhart1986}. In this stage, gradient computations are performed by applying the chain rule. Optimizers continue to update the parameters at each iteration as the data is processed and the model is gradually improved. Optimizers are typically categorized into two groups based on the type of information they use: first-order (e.g., gradients) or second-order (e.g., Hessians). Although second-order optimizers such as Newton's method and quasi-Newton methods are also available in the literature \cite{boyd2004,broyden1970,fletcher1970,goldfarb1970,shanno1970,nocedal2006,liu1989}, they are computationally expensive for deep learning. Therefore, first-order optimizers have primarily been the focus, mainly due to their lower computational cost. Nevertheless, research into reducing the computational cost of second-order optimizers is still ongoing \cite{martens2010,martens2015,gupta2018,yao2021}.

In today's deep learning environment, there are numerous well-known or recently introduced optimizers developed for use in training deep learning models. This paper aims to examine these optimizers and highlight their distinctive aspects. In the paper, each optimizer is discussed separately, the update rules of the optimizers are presented, and the variables in the equations are explained with their corresponding notation. The trainable parameters are simply referred to as the weights ($w$) in the rest of this paper. The weights are elements of $\mathbb{R}^n$, the $n$-dimensional space of real numbers. $L(w)$ denotes the loss function, which is a function of the weights. $\nabla_w$ is the gradient operator with respect to the weights. The subscript $t$ indicates the update iteration or the update time step in training at which the forward pass and backpropagation occur for each update. At the end of the $t$th update iteration, the weights are updated and then used in the $(t+1)$th iteration. $\eta$ is the learning rate, also referred to as the global learning rate in this paper, which is used to control the size of the update steps during optimization. It is a small positive number, the typical value of which varies depending on the optimizer used, with typical values such as 0.1, 0.01, or 0.001. Note that the learning rate can either be maintained as a fixed value throughout training or adjusted in subsequent iterations using a learning rate schedule. It can be defined as a corresponding element of a set for each $t$th update iteration in the optimization process, i.e., $\{\eta_t\}_{t=1}^T$. In addition to these, other variables are defined at their first occurrence in the paper. To prevent redundancy, variables and concepts explained in previous optimizer sections will not be discussed again in the following optimizers.
The review is further supported by illustrative trajectory visualizations. These visualizations are intended to make the update mechanisms easier to interpret, rather than to rank optimizers for general deep learning performance. To provide a modest task-level anchor for these illustrations, the paper additionally reports small-scale transformer experiments under matched training budgets.

In accordance with the historical order of optimizers, those to be discussed in this paper are listed in Table~\ref{tab:optimizers} along with their introduction years and corresponding references.

\begin{table}[htbp]
\renewcommand{\arraystretch}{1.2}
\caption{Chronological list of deep learning optimizers with related references}
\begin{center}
\begin{tabular}{|>{\centering\arraybackslash}p{2.4cm}|>{\centering\arraybackslash}p{2.4cm}|>{\centering\arraybackslash}p{2.4cm}|}
\hline
\textbf{Optimizer}&\textbf{Year Introduced}&\textbf{Related References} \\
\hline
SGD & 1951 & \cite{cauchy1847,lemarechal2012,ruder2016,robbins1951,bottou2010} \\ \hline
Momentum & 1964 & \cite{polyak1964,qian1999} \\ \hline
NAG & 1983 & \cite{nesterov1983,sutskever2013} \\ \hline
Adagrad & 2011 & \cite{duchi2011} \\ \hline
RMSprop & 2012 & \cite{hinton2012} \\ \hline
Adadelta & 2012 & \cite{zeiler2012} \\ \hline
Adam & 2014 & \cite{kingma2014} \\ \hline
Adamax & 2015 & \cite{kingma2014} \\ \hline
Nadam & 2016 & \cite{dozat2016} \\ \hline
AdamW & 2017 & \cite{loshchilov2019} \\ \hline
Adafactor & 2018 & \cite{shazeer2018} \\ \hline
AMSgrad & 2018 & \cite{reddi2018} \\ \hline
Radam & 2019 & \cite{liu2020} \\ \hline
LAMB & 2019 & \cite{you2017,you2020} \\ \hline
Lookahead & 2019 & \cite{zhang2019lookahead} \\ \hline
Adabelief & 2020 & \cite{zhuang2020} \\ \hline
SAM & 2020 & \cite{foret2021} \\ \hline
ASAM & 2021 & \cite{kwon2021} \\ \hline
Ranger21 & 2021 & \cite{wright2021} \\ \hline
Adan & 2022 & \cite{xie2024} \\ \hline
Lion & 2023 & \cite{chen2023,rong2025} \\ \hline
Sophia & 2023 & \cite{liu2023} \\ \hline
Muon & 2024 & \cite{jordan2024,liu2025} \\ \hline
NorMuon & 2025 & \cite{li2025normuon} \\ \hline
\ScheduleFreePlus & 2026 & \cite{defazio2024schedulefree, defazio2026schedulefreeplus}  \\
\hline
\end{tabular}
\label{tab:optimizers}
\end{center}
\end{table}

\section{Deep Learning Optimizers}
In the following subsections, optimizers are introduced using their full names. For certain optimizers whose original papers did not specify the full name, their original abbreviated titles are retained. Furthermore, certain optimizer subsections also include references to related optimizers and methods.

\subsection{Stochastic Gradient Descent}
Gradient-based methods trace their origins to Cauchy's 1847 paper, which aimed to solve systems of simultaneous equations \cite{cauchy1847,lemarechal2012}. The paper presents a mathematical procedure for minimizing a continuous, non-negative multivariable function to zero. Gradient descent (GD) is the main optimization algorithm in machine learning and deep learning, which minimizes the loss function by utilizing its gradient with respect to the weights. The gradient gives the slope of the loss function for the current weights, indicating the direction of the steepest ascent of the function. The initial weight values of a deep learning model can be chosen randomly or determined according to certain methods such as Glorot initialization and He initialization. The GD optimizer updates the weights in the opposite direction of the gradient, using the entire dataset at once. Since all the data is processed for each update, training the model can be time-consuming, particularly with large datasets. GD is also called Batch gradient descent (BGD) or Vanilla gradient descent (Vanilla GD) \cite{ruder2016}. It can differ depending on the amount of data used in each update iteration.

Stochastic gradient descent (SGD) is a variant of GD that updates the weights using only a single data sample at a time \cite{robbins1951,bottou2010}. The stochasticity arises from randomly selecting each data sample as input. Although SGD increases the variance in the loss value, it leads to a reduction in training time. In addition, it prevents the algorithm from getting stuck in a local minimum, allowing it to move towards other local minima or the global minimum. As another variant, a small subset of data samples (i.e., a mini-batch) can also be used in each update, which is referred to as Mini-batch gradient descent (Mini-batch GD). The mini-batch size can be chosen depending on the dataset size, such as 64, 128, or 256 samples. This enables more balanced steps to improve model performance by reducing the variance. For large datasets, the use of mini-batches is widespread not only in this context but also in almost all optimizers. Therefore, it would be more accurate to regard it as a technique incorporated into SGD or other optimizers, rather than a separate optimizer. BGD, SGD, and Mini-batch GD follow the same update rule, regardless of the amount of data used in each update iteration. Considering the loss function of weights $L(w)$, the SGD optimizer iteratively changes the weights at the $t$th update iteration ($w_t$) to obtain the weights at the $(t+1)$th update iteration ($w_{t+1}$) as
\begin{equation}
w_{t+1} = w_t - \eta \nabla_w L(w_t),
\label{eq:sgd}
\end{equation}
where the learning rate $\eta$ is a key hyperparameter that helps adjust the rate of the optimization process and has a significant impact on model performance. A learning rate that is too small may significantly slow down the convergence of the algorithm, while an excessively large learning rate may prevent it from converging to the optimal solution. Therefore, selecting an appropriate learning rate is crucial for successful optimization. In practice, the learning rate is gradually reduced over time from its initial value as training progresses. $\nabla_w L(w_t)$ is the gradient of the loss function with respect to the weights at the $t$th update iteration (i.e., the update time step $t$). Multiplying the learning rate by the gradient determines the amount of change in the weights. The negative sign preceding this multiplication ensures movement in the direction of the steepest descent of the loss function.

\subsection{Momentum}
SGD updates the weights based on the current gradients. It converges gradually, often exhibiting oscillations and high variance due to its stochastic nature, while searching for the optimal solution. This may cause the algorithm to focus especially around steep local minima and waste unnecessary time. In order to overcome these problems, SGD with momentum, often referred to simply as Momentum, is applied to deep learning as an extended version of SGD \cite{rumelhart1986, polyak1964,qian1999}. Momentum updates the weights using a new variable, 'velocity', which incorporates both the current and past gradients. The name Momentum originates from a physical analogy, where the velocity corresponds to the momentum of a particle with unit mass. The velocity helps to reduce the loss value smoothly during training. This facilitates faster convergence of the algorithm with reduced oscillations towards the optimal solution. In this case, the velocity and Momentum update rules can be expressed as
\begin{equation}
{\mathrm{v}}_t = \gamma {\mathrm{v}}_{t-1} - \eta \nabla_w L(w_t),
\label{eq:mom_v}
\end{equation}
\begin{equation}
w_{t+1} = w_t + {\mathrm{v}}_t,
\label{eq:mom_w}
\end{equation}
where ${\mathrm{v}}_t$ is the velocity at the $t$th update iteration, and its initial value is usually assigned as zero in the absence of prior gradient information. $\gamma$ is the momentum parameter, representing an exponential decay factor ranging from 0 to 1, typically set to 0.9. A momentum parameter close to 1 increases the influence of momentum on the weight updates. When the current gradient aligns with previous gradients, the weights are updated by larger amounts.

\subsection{Nesterov Accelerated Gradient}
Another momentum-based technique is introduced as Nesterov accelerated gradient (NAG) algorithm to improve SGD \cite{nesterov1983,sutskever2013}. It aims to estimate the next value of the weights and update them accordingly. This provides a way for looking ahead and considering the idea that weight updates can be performed more accurately. Theoretically, NAG outperforms gradient descent in terms of the rate of convergence of the excess error, improving from $O(1/T)$ to $O(1/T^2)$, where $T$ denotes the number of iterations. The application of this algorithm results in less oscillation and faster convergence to the minima compared to SGD. Additionally, due to its anticipatory nature, NAG is expected to be more responsive to changes in the direction of the gradient than Momentum. By replacing $w_t$ with $w_t + \gamma {\mathrm{v}}_{t-1}$ in the gradient computation, the velocity update rule changes and then the NAG optimizer updates the weights as
\begin{equation}
{\mathrm{v}}_t = \gamma {\mathrm{v}}_{t-1} - \eta \nabla_w L(w_t + \gamma {\mathrm{v}}_{t-1}),
\label{eq:nag_v}
\end{equation}
\begin{equation}
w_{t+1} = w_t + {\mathrm{v}}_t.
\label{eq:nag_w}
\end{equation}

Following gradient descent and momentum-based optimizers, the development of adaptive learning rate methods represents the next step in this progression. Note that, for the sake of simplicity in the equations presented throughout the remainder of this paper, the gradient at the $t$th update iteration is denoted as follows
\begin{equation}
g_t = \nabla_w L(w_t).
\label{eq:g}
\end{equation}

\subsection{Adaptive Gradient Algorithm}
SGD can be limited in learning features that are highly informative and discriminative but rarely appear in the dataset. To address this limitation, adaptive optimizers are commonly proposed in the literature. Adaptive gradient algorithm (Adagrad) is one of the pioneering optimizers in gradient-based learning that introduced the concept of adaptive gradient, i.e.\ adaptive learning rate \cite{duchi2011}. Adagrad as an adaptive optimizer dynamically changes the learning rate for each weight while applying gradient descent. It normalizes the learning rate of each weight using the past squared gradients of that weight. Thus, some weights that are updated in relatively small amounts have a higher learning rate than others.

First, defining $g_t^2$ as the element-wise square of $g_t$, the accumulation of squared gradients up to (i.e., including) the $t$th update iteration is written as
\begin{equation}
a_t = \sum_{j=1}^t g_j^2 = a_{t-1} + g_t^2.
\label{eq:adagrad_a}
\end{equation}

Then, by using the square root of $a_t$ as the normalization term for the learning rate in Adagrad, the next weights are computed as
\begin{equation}
w_{t+1} = w_t - \frac{\eta}{\sqrt{a_t + \epsilon}} g_t,
\label{eq:adagrad}
\end{equation}
where $\epsilon$ is the smoothing term avoiding division by zero, a small constant typically set to $10^{-7}$ or $10^{-8}$. Note that all mathematical operations on vectors are applied element-wise to calculate updates for each individual weight.

\subsection{Root Mean Square Propagation}
The accumulation of squared gradients, and consequently the normalization term, continuously increases throughout training in Adagrad. This causes the learning rate to become very small over time, thereby preventing the model from learning effectively. Root mean square propagation (RMSprop) as an optimizer can be regarded as an effective approach to addressing this issue \cite{hinton2012}. RMSprop modifies Adagrad by employing an exponentially decaying average of past squared gradients up to the $t$th update iteration. It is stated as
\begin{equation}
b_t = \beta b_{t-1} + (1 - \beta) g_t^2,
\label{eq:rms_b}
\end{equation}
where $\beta$ is the decay rate, typically set to 0.9. The square root of $b_t$ corresponds to a decaying root mean square of the gradients, from which the name RMSprop is derived. By employing this root mean square in the normalization term, the influence of past gradients is diminished, enabling faster convergence on non-convex functions compared to Adagrad. Although $b_t$ is computed differently from $a_t$ in Adagrad, the RMSprop update rule is expressed in a similar form as
\begin{equation}
w_{t+1} = w_t - \frac{\eta}{\sqrt{b_t + \epsilon}} g_t.
\label{eq:rms}
\end{equation}

This update rule prevents the learning rate from decreasing too quickly. It is suitable for non-stationary cases where the data distribution changes over time.

\subsection{Adadelta}
Similar to RMSprop, Adadelta is a variant of Adagrad that aims to overcome the decreasing learning rate problem \cite{zeiler2012}. It updates the weights without the need to manually specify a global learning rate. It is worth noting that Adadelta is used as the name of the algorithm in the original paper, rather than an abbreviation. However, it can be readily inferred that the term 'delta' originates from the weight updates $\Delta w$, which represent the changes applied to the weights. Accordingly, the update rule can be expressed in its most general form as
\begin{equation}
w_{t+1} = w_t + \Delta w_t.
\label{eq:adadelta_gen}
\end{equation}

In Adadelta, $c_t$ is defined as the mean square of weight updates up to the $t$th update iteration, similar to $b_t$, as
\begin{equation}
c_t = \beta c_{t-1} + (1 - \beta) \Delta w_t^2.
\label{eq:adadelta_c}
\end{equation}

The weight update at the $t$th update iteration ($\Delta w_t$) is computed by utilizing $c_{t-1}$ and $b_t$ as follows
\begin{equation}
\Delta w_t = -\frac{\sqrt{c_{t-1} + \epsilon}}{\sqrt{b_t + \epsilon}} g_t.
\label{eq:adadelta_delta}
\end{equation}

The inclusion of $\epsilon$ in the numerator ensures that updates occur both at the beginning of the optimization process and in cases where the weight updates become very small. In this case, the Adadelta update rule can be written as
\begin{equation}
w_{t+1} = w_t - \frac{\sqrt{c_{t-1} + \epsilon}}{\sqrt{b_t + \epsilon}} g_t.
\label{eq:adadelta}
\end{equation}

Here, the learning rate is replaced by the root mean square of weight updates. Thus, the learning rate is not used during the training of the model.

\subsection{Adaptive Moment Estimation}
Momentum-based optimizers (Momentum and NAG) and adaptive optimizers (Adagrad, RMSprop, and Adadelta) constitute the algorithms discussed thus far, each aiming to improve the performance of SGD through distinct strategies. Adaptive moment estimation (Adam) is another adaptive optimization algorithm that aims to take advantage of these optimizers by bringing them together \cite{kingma2014}. It uses the first moment estimate of gradients ($m_t$) and the second moment estimate of gradients ($v_t$) at the $t$th update iteration, expressed as
\begin{equation}
\begin{aligned}
m_t = \beta_1 m_{t-1} + (1 - \beta_1) g_t, \\
v_t = \beta_2 v_{t-1} + (1 - \beta_2) g_t^2.
\end{aligned}
\end{equation}

$\beta_1$ is the first decay rate, and $\beta_2$ is the second decay rate, where $\beta_1, \beta_2 \in [0,1)$. These rates are typically set to 0.9 and 0.999, respectively. Initializing the moment estimates as zero vectors introduces a bias towards zero, particularly during the initial update iterations. In order to avoid bias towards zero, $m_t$ and $v_t$ are corrected as
\begin{equation}
\hat{m}_t = \frac{m_t}{1 - \beta_1^t}, \quad \hat{v}_t = \frac{v_t}{1 - \beta_2^t},
\label{eq:adam_bias}
\end{equation}
where $\hat{m}_t$ and $\hat{v}_t$ are the bias-corrected moment estimates. $\beta_1^t$ and $\beta_2^t$ represent $\beta_1$ and $\beta_2$ to the power $t$, respectively. Finally, the Adam optimizer updates the weights as
\begin{equation}
w_{t+1} = w_t - \frac{\eta}{\sqrt{\hat{v}_t + \epsilon}} \hat{m}_t.
\label{eq:adam}
\end{equation}

In this way, Adam adaptively adjusts the learning rate for each weight using first-order gradients. It converges to the optimal solution rapidly in most cases. Also, the hyperparameters in Adam generally require minimal tuning. In the original paper, the default values for $\eta$ and $\epsilon$ are given as 0.001 and $10^{-8}$, respectively.

Note that for Adam and other optimizers, the learning rate may also be defined depending on $t$ as $\eta_t$ 
instead of a constant $\eta$. 

\subsection{Adamax}
In Adam's update, the $L_2$ norm of gradients is used as a scaling factor to determine the amount of change in the weights. Instead, the Adamax optimizer, introduced by the same authors as a variant of Adam to achieve more stability, proposes scaling with the $L_p$ norm where $p$ goes to infinity \cite{kingma2014}. The second moment estimate of gradients $v_t$ is changed according to $p$ as
\begin{equation}
v_t = \beta_2^p v_{t-1} + (1 - \beta_2^p) |g_t|^p,
\end{equation}
and the exponentially weighted infinity norm ($u_t$) is defined as
\begin{equation}
u_t = \lim_{p \to \infty} (v_t)^{1/p}.
\label{eq_ut}
\end{equation}

Upon applying the manipulations by substituting $v_t$ in (\ref{eq_ut}), $u_t$ is expressed in the following simple form as
\begin{equation}
u_t = \max(\beta_2 u_{t-1}, |g_t|),
\end{equation}
where the initial value of $u_t$ is zero, $u_0 = 0$. It can be inferred that the 'max' in the name Adamax is attributed to the use of the maximum operation. As the maximum operation does not exhibit a bias towards zero, $u_t$ does not require a bias correction. By using $u_t$, the Adamax update rule becomes
\begin{equation}
w_{t+1} = w_t - \frac{\eta}{u_t} \hat{m}_t.
\label{eq:adamax}
\end{equation}

\subsection{Nesterov Accelerated Adam}
Although Adam is an efficient and popular optimizer, various studies are being carried out to achieve better results. Adam's adaptive moment estimation and NAG's accelerated gradient methods are combined in Nesterov accelerated Adam (Nadam) \cite{dozat2016}. In order to obtain faster convergence, it proposes to adapt the look ahead approach into Adam. In Nadam, the first moment estimate used in Adam is first scaled and then extended with the Nesterov estimate as follows
\begin{equation}
\hat{m}_{N,t} = \frac{\beta_1}{1 - \beta_1^{t+1}} m_t + \frac{(1 - \beta_1)}{1 - \beta_1^t} g_t,
\end{equation}
where $\hat{m}_{N,t}$ can be referred to as the bias-corrected first moment estimate in Nadam. The Adam update rule is rearranged by replacing $\hat{m}_t$ with $\hat{m}_{N,t}$, and the Nadam update rule is stated as
\begin{equation}
w_{t+1} = w_t - \frac{\eta_t}{\sqrt{\hat{v}_t + \epsilon}} \hat{m}_{N,t}.
\label{eq:nadam}
\end{equation}

With this formulation, the Nadam optimizer is anticipated to achieve improved learning stability and convergence speed, while introducing a slight increase in computational load. According to the experiments conducted in \cite{dozat2016}, Nadam achieves its best performance with the hyperparameter values $\beta_1 = 0.975$, $\beta_2 = 0.999$, and $\epsilon = 10^{-8}$.

Note that the first decay rate in Nadam may also be defined depending on $t$ as $\beta_{1t}$ instead of a constant $\beta_{1}$ as in the original paper. 

\subsection{Adam with Decoupled Weight Decay}
Adam with decoupled weight decay (AdamW) is another variant of Adam, proposing an alternative weight decay approach \cite{loshchilov2019}. The Adam optimizer updates the weights by using the gradients of the loss function. AdamW argues that the weight update in Adam, which is usually implemented by $L_2$ regularization, does not constitute true weight decay. Instead, AdamW decouples weight decay from gradient based updating. This provides an independent weight decay regularization and helps to improve generalization performance. According to AdamW, the updates are performed as follows
\begin{equation}
w_{t+1} = w_t - \eta_t \left( \frac{\eta}{\sqrt{\hat{v}_t + \epsilon}} \hat{m}_t + \lambda w_t \right),
\label{eq:adamw}
\end{equation}
where $\lambda$ is the weight decay rate that applies the same regularization rate to all weights. The term $\eta_t$ serves as a scaling factor (such as the learning rate at the $t$th update iteration) for the user-defined scheduling of both $\eta$ and $\lambda$. Here, $\eta$ and $\eta_t$ are decoupled to obtain the actual learning rate at the $t$th update iteration.

Additionally, the same paper suggests employing warm restarts to enhance AdamW's anytime performance. The proposed AdamW with warm restarts (AdamWR) algorithm is implemented by applying $\eta_t$ that decays according to a cosine annealing. $\eta_t$ can be expressed as
\begin{equation}
\eta_t = \eta_{\min} + 0.5 (\eta_{\max} - \eta_{\min}) (1 + \cos(\pi T_{\text{cur}} / T_i))
\end{equation}
and for $\eta_{\min}=0$ and $\eta_{\max}=1$, i.e., for values within the range $[0, 1]$, as follows
\begin{equation}
\eta_t = 0.5 + 0.5 \cos(\pi T_{\text{cur}} / T_i).
\label{eq:adamwr}
\end{equation}

The index $i$ denotes the order of the warm restart cycles. $T_i$ is the total number of update iterations in the $i$th cycle, and $T_{\text{cur}}$ is the current iteration within the $i$th cycle. When $T_{\text{cur}} = T_i$, the current cycle is completed, $\eta_t$ is reset, and a new cycle begins. Note that $\eta_t$ can be used in place of the learning rate both here and in other optimizers, without being decoupled from $\eta$.

\subsection{Adafactor}
During training, adaptive optimizers require additional memory to store variables such as the first and second moment estimates, as used in Adam, in addition to the model weights. As neural networks scale and the number of weights increases, memory requirements become a significant constraint. Adafactor is a variant of Adam, but unlike other algorithms, it aims to make resource usage more efficient \cite{shazeer2018}. The Adafactor optimizer uses a number of techniques to reduce memory usage and increase computational efficiency.

Reducing memory usage is achieved by representing matrix-valued variables using the moving averages of their row and column sums. Assuming that a subset of the model weights is arranged by a matrix $W$, a matrix $V$ of the same size is needed for the second moment estimate of gradients. Hence, if $W \in \mathbb{R}^{n \times m}$, then $V \in \mathbb{R}^{n \times m}$ as well. The second moment estimate matrix can be approximated as the outer product of the row and column vectors $R \in \mathbb{R}^{n \times 1}$ and $C \in \mathbb{R}^{1 \times m}$, respectively, $V \approx R C$. In this case, the matrix $V$ requires memory of size $n + m$ instead of $n \times m$. Considering matrix form, the bias-corrected moment estimate matrix at the $t$th update iteration ($\hat{V}_t$) and the factors ($R_t$, $C_t$) are written as
\begin{equation}
\begin{gathered}
R_t = \hat{\beta}_{2t} R_{t-1} + (1 - \hat{\beta}_{2t}) (G_t^2 + \epsilon_1 \mathbf{1}_n \mathbf{1}_m^\top) \mathbf{1}_m, \\
C_t = \hat{\beta}_{2t} C_{t-1} + (1 - \hat{\beta}_{2t}) \mathbf{1}_n^\top (G_t^2 + \epsilon_1 \mathbf{1}_n \mathbf{1}_m^\top), \\
\hat{V}_t = \frac{R_t C_t}{\mathbf{1}_n^\top R_t},
\label{eq:adafactor_v}
\end{gathered}
\end{equation}
where $\hat{\beta}_{2t}$ denotes the corrected second decay rate. $\epsilon_1$ is the first regularization constant. $\mathbf{1}_n$ and $\mathbf{1}_m$ represent column vectors of ones of size $n$ and $m$, respectively. $\top$ is the transpose operator.

To further reduce memory usage, Adafactor proposes removing the first moment estimate used in Adam ($\beta_1 = 0$). However, this removal can lead to large updates and training instability. To address this issue, Adafactor introduces an increasing decay rate schedule and update clipping. In this context, $\hat{\beta}_{2t}$ is defined as
\begin{equation}
\hat{\beta}_{2t} = 1 - \frac{1}{t^e}, \quad t \geq 1,
\label{eq:adafactor_beta}
\end{equation}
where $e > 0$ is a scalar that controls the rate of increase. Also, the unscaled update $U_t$ and the clipped unscaled update $\hat{U}_t$ are defined as
\begin{equation}
U_t = \frac{G_t}{\sqrt{\hat{V}_t}},
\end{equation}
\begin{equation}
\hat{U}_t = \frac{U_t}{\max(1, \text{RMS}(U_t)/\tau_c)},
\label{eq:adafactor_clip}
\end{equation}
where $\text{RMS}(\cdot)$ is the root mean square operator, and $\tau_c$ is the clipping threshold. Thus, when $\text{RMS}(U_t)$ exceeds $\tau_c$, the updates are scaled down to prevent large updates.

At the same time, Adafactor dynamically adjusts the learning rate by scaling the update magnitude relative to the scale of the weights. Instead of using a fixed learning rate, it proposes determining the learning rate via a relative step size ($q_t$). The learning rate is expressed as
\begin{equation}
\eta_t = \max(\epsilon_2, \text{RMS}(w_t)) q_t,
\label{eq:adafactor_eta}
\end{equation}
where $\epsilon_2$ is the second regularization constant. Eventually, by incorporating $\eta_t$ and $\hat{U}_t$ into the computation, the Adafactor update rule is written as
\begin{equation}
w_{t+1} = w_t - \eta_t \hat{U}_t.
\label{eq:adafactor}
\end{equation}

The hyperparameters for Adafactor are proposed in the original paper as $\epsilon_1 = 10^{-30}$, $\epsilon_2 = 10^{-3}$, $\tau_c = 1$, $q_t = \min(10^{-2}, 1/\sqrt{t})$, and $\hat{\beta}_{2t} = 1 - t^{-0.8}$. Given these hyperparameters and the proposed techniques, Adafactor performs effectively, particularly on large-scale models.

\subsection{AMSgrad}
In some cases, Adam may not converge to the optimal solution. Due to factors such as sparse gradients and high $\beta_2$ values, the second moment estimate $v_t$ may decrease over time, which can lead to an excessively large learning rate. In this case, Adam may fail to minimize the loss function or to converge to an optimal solution. It is considered that the source of this issue is the way the second moment estimate $v_t$ is updated. Adam updates $v_t$ by computing the decaying average of the past squared gradients. To mitigate the convergence issue, AMSgrad maintains the maximum of past second moment estimates \cite{reddi2018}. Thus, $v_t$ in AMSgrad decays more slowly than in Adam, improving stability during optimization. Its mathematical definition is modified as follows
\begin{equation}
\hat{v}_t = \max(\hat{v}_{t-1}, v_t)
\label{eq:ams_max}
\end{equation}
and by removing the debiasing step for the sake of simplicity, the AMSgrad update rule is implemented as
\begin{equation}
w_{t+1} = w_t - \frac{\eta_t}{\sqrt{\hat{v}_t + \epsilon}} m_t.
\label{eq:amsgrad}
\end{equation}

In the same paper, AdamNC is proposed as an optimizer that utilizes time-varying decay rates ($\beta_{1t}$, $\beta_{2t}$) instead of Adam's constant decay rates $\beta_1$ and $\beta_2$. Specifically, $\beta_{2t}$ increases over time according to an increasing schedule, which allows the second moment estimate to be calculated in a more controlled manner. In AdamNC, the weight updates are performed according to Adam's update rule using the following equations.
\begin{equation}
\begin{gathered}
m_t = \beta_{1t} m_{t-1} + (1 - \beta_{1t}) g_t, \\
v_t = \beta_{2t} v_{t-1} + (1 - \beta_{2t}) g_t^2.
\end{gathered}
\end{equation}

Here, $\beta_{2t}$ can be selected as $\beta_{2t} = 1 - 1/t$, similar to the approach used in Adafactor.

\subsection{Rectified Adam}
In Adam, the learning rate changes adaptively, but the variance of this change is quite large in the early stages of optimization. Rectified Adam (Radam), as a variant of Adam, aims to solve the variance problem of learning rate \cite{liu2020}. Radam controls the variance by adding a rectification term, which justifies the learning rate warmup heuristic. This rectification term makes the optimization more stable in the early stages of the training process.

Firstly, Radam defines the degrees of freedom ($\rho_t$) for the analytic form of variance as
\begin{equation}
\rho_t = \frac{2}{1 - \beta_2} - 1 - \frac{2 t \beta_2^t}{1 - \beta_2^t},
\label{eq:radam_rho}
\end{equation}
and the variance rectification term ($r_t$) as
\begin{equation}
r_t = \sqrt{\frac{(\rho_t - 4)(\rho_t - 2) \rho_\infty}{(\rho_\infty - 4)(\rho_\infty - 2) \rho_t}},
\label{eq:radam_r}
\end{equation}
where $\rho_\infty = 2/(1 - \beta_2) - 1$.

Then, if the variance is tractable, i.e., $\rho_t > 4$, Radam updates the weights using the variance rectification term as
\begin{equation}
w_{t+1} = w_t - \frac{\eta_t r_t}{\sqrt{\hat{v}_t}} \hat{m}_t = w_t - \eta_t r_t l_t \hat{m}_t,
\label{eq:radam_rect}
\end{equation}
where $l_t$ is defined as the adaptive learning rate in the original paper, $l_t = 1/\sqrt{\hat{v}_t}$.

Otherwise, the weights are updated using the first moment estimate without adaptation as follows
\begin{equation}
w_{t+1} = w_t - \eta_t \hat{m}_t.
\label{eq:radam_no}
\end{equation}

Thus, two distinct update strategies are integrated into a single update rule to reduce excessive volatility. In the Radam optimizer, the same hyperparameter settings can be used as in Adam.

\subsection{Layerwise Adaptive Moments Optimizer for Batch Training}
The computational cost of training large-scale deep neural networks is quite high. To address this issue, the Layerwise adaptive rate scaling (LARS) optimizer proposes the use of layerwise adaptive learning rates for training networks on massive datasets with large batch sizes \cite{you2017}. However, although LARS performs well in certain tasks such as residual networks, it does not consistently achieve comparable performance across all tasks. This shortcoming of LARS is addressed by an alternative layerwise optimizer. Layerwise adaptive moments optimizer for batch training (LAMB) is a layerwise variant of Adam that develops a new adaptation strategy to reduce the cost and increase training speed \cite{you2020}. LAMB provides this adaptivity through both per-dimension normalization using the square root of the second moment, and layerwise normalization. Although the learning rate of each weight is adjusted separately as in Adam, it allows each layer to learn at different speeds by adjusting the learning rates of the weights for each layer.

Defining a scaling function $\phi : \mathbb{R}^+ \to \mathbb{R}^+$, the LAMB update rule for the $i$th layer is expressed as
\begin{equation}
w_{t+1}^i = w_t^i - \eta_t \frac{\phi(\|w_t^i\|)}{\|x_t^i + \lambda w_t^i\|} (x_t^i + \lambda w_t^i),
\label{eq:lamb}
\end{equation}
where the superscript $i$ in the variables indicates that the corresponding variable belongs to the $i$th layer. $x_t^i$ is a ratio computed as $x_t^i = \hat{m}_t / \sqrt{\hat{v}_t + \epsilon}$, and $\|\cdot\|$ denotes $L_2$ norm of a vector. In the original paper, the following scaling function was observed to perform well
\begin{equation}
\phi(z) = \min(\max(z, \gamma_l), \gamma_u),
\label{eq:lamb_phi}
\end{equation}
where $\gamma_l$ and $\gamma_u$ represent the lower and upper limits of the function, respectively.

The update rule of LAMB improves the optimizer's convergence towards the target, especially for large batch sizes such as 32K. In the same paper, Nesterov LAMB (N-LAMB) is introduced as a variant of LAMB that replaces the first moment estimate with Nesterov momentum. Additionally, NN-LAMB is presented, in which Nesterov momentum is applied to both the first and second moment estimates. The performance of N-LAMB and NN-LAMB is comparable to that of LAMB.

\subsection{Lookahead}
Many deep learning optimization algorithms are based on momentum or adaptive learning-rate approaches. Unlike these approaches, the Lookahead optimizer proposes iteratively updating two sets of weights, slow weights and fast weights, during optimization \cite{zhang2019lookahead}. It uses a standard optimization algorithm (e.g., SGD or Adam) as an inner optimizer and a synchronization step as a lookahead mechanism. The slow weights are synchronized with the fast weights obtained from the inner optimizer every $k$ inner iterations. Thus, the optimization direction is guided and optimizer performance can be improved at a low cost. In Lookahead, for a loss function ($L$) and a sample mini-batch of data ($d$), the fast weights ($\theta$) are updated according to the inner optimizer ($A$) used as
\begin{equation}
\theta_{t,i} = \theta_{t,i-1} + A(L, \theta_{t,i-1}, d),
\label{eq:lookahead_fast}
\end{equation}
where $i$ is the inner loop variable, $i=1,2,\dots,k$. Then, the slow weights ($\phi$) are updated once every $k$ inner iterations as an exponential moving average of the $k$th fast weights
\begin{equation}
\phi_t = \phi_{t-1} + \alpha (\theta_{t,k} - \phi_{t-1}),
\label{eq:lookahead_slow}
\end{equation}
where $t$ is the outer loop variable, $t=1,2,\dots$ (such as the update time step $t$). The parameter $\alpha$ denotes the slow weights learning rate. After each slow weights update, the fast weights are synchronized before the inner iterations begin, $\theta_{t,0} = \phi_{t-1}$. Thus, with successive inner and outer iterations performed, optimization is achieved as directed by the slow weights. The slow weights exhibit more balanced changes with lower variance compared to the fast weights. By updating the weights ($w_t$) of the model with the slow weight values, the Lookahead optimizer enhances the stability of the training process.

\subsection{Adabelief}
Adaptive optimizers like Adam converge faster than SGD, but their stability and generalization performance may not be good enough for large datasets. Adabelief proposes a modification to the Adam optimizer as a way to improve training stability and model generalization \cite{zhuang2020}. Adabelief approaches the amount of change to be applied to the weights through the concept of 'belief'. In Adabelief, unlike Adam, a new variable $s_t$ is defined as the exponential moving average of $(g_t - m_t)^2$, where $g_t - m_t$ represents the deviation of the gradient. It is considered that $m_t$ is the prediction of the gradient, and the inverse square root of $s_t$, $1/\sqrt{s_t}$, is the 'belief' in the algorithm. A small difference between $g_t$ and $m_t$ means that there is a strong belief and hence a large change is carried out in the weights, whereas in the case of weak belief the change is small.

Based on this approach, the Adabelief optimizer computes $s_t$ and its bias correction $\hat{s}_t$ instead of $v_t$ and $\hat{v}_t$, respectively, as follows
\begin{equation}
\begin{gathered}
s_t = \beta_2 s_{t-1} + (1 - \beta_2) (g_t - m_t)^2 + \epsilon, \\
\hat{s}_t = \frac{s_t}{1 - \beta_2^t},
\end{gathered}
\end{equation}
and the Adabelief update rule is expressed as
\begin{equation}
w_{t+1} = w_t - \frac{\eta}{\sqrt{\hat{s}_t + \epsilon}} \hat{m}_t.
\label{eq:adabelief}
\end{equation}

Adabelief employs the same hyperparameters as Adam and can be used with the same default values.

\subsection{Sharpness-Aware Minimization}
In deep learning, optimizers typically perform optimization by minimizing the loss value. Sharpness-aware minimization (SAM) is an alternative optimizer that aims to improve generalization by exploiting the geometry of the loss landscape \cite{foret2021}. This optimizer focuses on minimizing the loss sharpness, which is the measure of how fast the loss changes in the neighborhood of the weights, as well as the loss value. It allows the model to avoid high-sharpness local minima and move towards low-sharpness minima, which increases the generalization performance. The loss sharpness is defined as
\begin{equation}
\max_{\|\epsilon\| \leq \sigma} L(w + \epsilon) - L(w),
\label{eq:sam_sharp}
\end{equation}
where $\sigma$ is the radius of maximization region as a non-negative hyperparameter, 0.05 by default. $\epsilon$ is the weight perturbation, maximizing the loss under the constraint $\sigma$. The weight perturbation at the update time step $t$ ($\epsilon_t$) is expressed and computed approximately as follows
\begin{equation}
\epsilon_t = \arg\max_{\|\epsilon\|_p \leq \sigma} L(w_t + \epsilon) \approx \sigma \frac{g_t}{\|g_t\|},
\label{eq:sam_epsilon}
\end{equation}
where $\|\cdot\|_p$ denotes the $L_p$ norm, and $p=2$ yields the optimal result. Since the optimizer aims to minimize both the loss and the sharpness, the problem becomes a min-max optimization problem. Using SGD as the base optimizer, SAM solves this problem and updates the weights as
\begin{equation}
w_{t+1} = w_t - \eta \nabla_w L(w_t + \epsilon_t).
\label{eq:sam}
\end{equation}

Here, the gradient is computed at the weights $w_t + \epsilon_t$ instead of the original weights $w_t$, and SGD is performed accordingly.

\subsection{Adaptive Sharpness-Aware Minimization}
In SAM, rescaling of weights preserves the loss value but alters the loss sharpness. Due to this sensitivity to weight rescaling, the loss sharpness definition may negatively impact generalization performance. Adaptive sharpness-aware minimization (ASAM) is a variant of the SAM optimizer that aims to further increase the generalization performance of the model \cite{kwon2021}. For this aim, ASAM introduces the concept of adaptive sharpness that satisfies the scale-invariant property. The adaptive sharpness is defined as
\begin{equation}
\max_{\|T_w^{-1} \epsilon\|_p \leq \sigma} L(w + \epsilon) - L(w),
\label{eq:asam_sharp}
\end{equation}
where $T_w$ is the normalization operator of $w$, and $T_w^{-1}$ denotes the inverse of $T_w$. Based on \eqref{eq:asam_sharp}, the weight perturbation at the update time step $t$ becomes
\begin{equation}
\epsilon_t = \sigma \frac{T_{w_t}^2 g_t}{\|T_{w_t} g_t\|}.
\label{eq:asam_epsilon}
\end{equation}

Eventually, as ASAM's solution to the min-max optimization problem, the update rule is stated as
\begin{equation}
w_{t+1} = w_t - \eta (\nabla_w L(w_t + \epsilon_t) + \lambda w_t).
\label{eq:asam}
\end{equation}

Note that the base optimizer in SAM and ASAM can be replaced by another optimizer such as Adam. In addition, adaptive sharpness is proposed in the original paper as a generalization assessment metric for neural networks.

\subsection{Ranger21}
Ranger is an optimizer introduced in 2019 that combines Radam and Lookahead optimization algorithms \cite{wright2019}. The Ranger21 optimizer is the evolved version of Ranger in 2021 \cite{wright2021}. Ranger21 consists of a main optimizer, AdamW, and eight components that are combined with it. These components, each containing a useful technique, are Adaptive gradient clipping, Gradient centralization, Positive-negative momentum, Linear learning rate warm-up, Explore-exploit learning rate schedule, Stable weight decay, Norm loss, and Lookahead. These techniques are considered to be orthogonal and synergistic in the context of optimization.

The optimizer begins with the gradient computation, and if $\|g_t\| / \max(\|w_t\|, \epsilon_c) > \tau_c$ then continues with Adaptive gradient clipping \cite{brock2021} as follows
\begin{equation}
g_t = \tau_c \frac{\max(\|w_t\|, \epsilon_c)}{\|g_t\|} g_t,
\label{eq:ranger_clip}
\end{equation}
where $\epsilon_c$ is the smoothing term used in clipping to prevent the freezing of zero-initialized weights. This process is performed on individual rows, not on a full layer. Adaptive gradient clipping limits extremely large gradients to make the optimizer more stable. Gradient centralization \cite{yong2020} is the next step applied here as
\begin{equation}
g_t = g_t - \text{mean}(g_t),
\label{eq:ranger_cent}
\end{equation}
which is used as a normalization technique to improve generalization. Then, proceed to Positive-negative momentum \cite{xie2021} that uses positive and negative weights for the current momentum estimate and the previous one, respectively. Formulation steps of AdaPNM, a variant of Positive-negative momentum, differ from Adam in some points and can be written as
\begin{equation}
\begin{gathered}
m_t = \beta_1^2 m_{t-2} + (1 - \beta_1^2) g_t, \\
\hat{m}_t = \frac{(1 + \beta_0) m_t - \beta_0 m_{t-1}}{1 - \beta_1^t}, \\
v_t = \beta_2 v_{t-1} + (1 - \beta_2) g_t^2, \\
v_{\max} = \max(v_t, v_{\max}), \\
\hat{v}_t = v_{\max} / (1 - \beta_2^t), \\
y_t = \frac{\hat{m}_t}{\sqrt{(1 + \beta_0)^2 + \beta_0^2}(\sqrt{\hat{v}_t + \epsilon})},
\label{eq:ranger_pnm}
\end{gathered}
\end{equation}
where $\beta_0$ is the zeroth decay rate, $v_{\max}$ is the second moment maximum estimate, and $y_t$ denotes the update vector. Note that other variable names are already defined in Adam. This technique adds noise to the gradient, which helps the optimizer escape saddle points and converge toward flatter minima. Thus, the technique aims to improve generalization.

Linear learning rate warm-up, as used in Ranger21, is an optimization technique that aims to avoid excessive step size during the first iterations \cite{ma2021}. It refers to starting the training with a low learning rate and gradually increasing it throughout the training. In Ranger21, to avoid an overly long warm-up schedule, the number of warm-up iterations ($t_{\text{wup}}$) is defined, by default, as 22\% of the total number of iterations ($t_{\max}$). Eventually, the warm-up behavior of the learning rate is stated as
\begin{equation}
\eta_t = \min\left(1, \max\left( \frac{1-\beta_2}{2} t, \frac{t}{t_{\text{wup}}}\right)\right) \eta.
\label{eq:ranger_warm}
\end{equation}

Explore-Exploit learning rate schedule is a scheme consisting of sequential Explore and Exploit stages for good generalization \cite{iyer2023}. In the first stage, Explore, the aim is to reach a wide minimum region by optimizing with a high learning rate. In the next stage, Exploit, the learning rate is gradually reduced towards zero and a search is made to find the bottom of this region. In Ranger21, the number of warm-down iterations ($t_{\text{wdown}}$) is defined, by default, as 28\% of $t_{\max}$ for the Exploit stage, and the warm-down behavior of the learning rate is expressed as
\begin{equation}
\eta_t = \min\left(1, \frac{t_{\max} - t}{t_{\text{wdown}}}\right) \eta.
\label{eq:ranger_wdwn}
\end{equation}

Ranger21 combines the linear learning rate warm-up and the Explore-Exploit learning rate schedule, as formulated in equations \eqref{eq:ranger_warm} and \eqref{eq:ranger_wdwn}. Following the Positive-negative momentum technique, Ranger21 proceeds with the next step as follows
\begin{equation}
\eta_t = \min\left(1, \max\left( \frac{1-\beta_2}{2} t, \frac{t}{t_{\text{wup}}}\right), \frac{t_{\max} - t}{t_{\text{wdown}}}\right) \eta.
\label{eq:ranger_eta}
\end{equation}

Stable weight decay regularization proposes applying an isotropic bias correction term to stabilize the weight decay \cite{xie2021stable}. The stable step size ($d_s$) to be applied to the weights for Stable weight decay is
\begin{equation}
d_s = -\frac{\eta_t}{\sqrt{\text{mean}(\hat{v}_t)}} \lambda w_{t-1},
\label{eq:ranger_stable}
\end{equation}
where $\text{mean}(\cdot)$ represents the layerwise average.

Norm loss is a soft-regularization method in which each weight vector is pushed to have a norm close to one \cite{georgiou2020}. It aims to address the issues of exploding or vanishing gradients. For this aim, the norm loss step size ($d_n$) is defined as
\begin{equation}
d_n = -\eta_t \lambda \left(1 - \frac{1}{\|w_{t-1}\|}\right) w_{t-1}.
\label{eq:ranger_norm}
\end{equation}

In Ranger21, Stable weight decay and Norm loss are combined, and the total step size at the update time step $t$ ($d_t$) becomes
\begin{equation}
d_t = \frac{\eta_t}{\sqrt{\text{mean}(\hat{v}_t)}} \lambda \left(1 - \frac{1}{\|w_{t-1}\|}\right) w_{t-1}.
\label{eq:ranger_d}
\end{equation}

Based on all these different components, the update rule of Ranger21 is expressed as
\begin{equation}
w_t = w_{t-1} - \eta_t y_t - \eta_t d_t.
\label{eq:ranger21}
\end{equation}

After this weight update, the Lookahead algorithm is added to the Ranger21 optimizer as the last step. For this purpose, Ranger21 considers the previous weights as the fast weights and updates them in the direction of slow weights once in every $k$ iterations, in accordance with the concept introduced in the Lookahead section. As a result of applying the Lookahead algorithm every $k$ iterations, the weights are updated as
\begin{equation}
l_{t/k} = \beta_{\text{la}} l_{t/k-1} + (1 - \beta_{\text{la}}) w_t,
\end{equation}
\begin{equation}
w_t = l_{t/k},
\label{eq:ranger_look}
\end{equation}
where $\beta_{\text{la}}$ is the lookahead decay rate, and $l_{t/k}$ indicates the lookahead (slow) weights.

The hyperparameters for Ranger21 are proposed in the original paper as $k=5$, $\lambda=10^{-4}$, $\tau_c=10^{-2}$, $\epsilon=10^{-8}$, $\epsilon_c=10^{-3}$, $\beta_0=0.9$, $\beta_1=0.9$, $\beta_2=0.999$, and $\beta_{\text{la}}=0.5$. With the inclusion of various hyperparameters and components, Ranger21 is expected to be more robust, despite having a more complex structure.

\subsection{Adaptive Nesterov Momentum}
Each deep learning network may require a different optimizer to perform well. Determining this empirically often makes the training process inefficient. Adaptive Nesterov momentum (Adan) is an optimizer designed to be suitable for various types of deep learning networks and to enhance the training process \cite{xie2024}. Adan adopts a modified approach to Nesterov momentum. Instead of computing the gradient at a forward (extrapolated) point, it estimates the momentum direction based on the current and previous gradients. This method, named Nesterov momentum estimation (NME), avoids additional computational cost. According to NME, Adan changes the first and second moment estimates of gradients in Adam as
\begin{equation}
\begin{gathered}
m_t = (1 - \beta_1) m_{t-1} + \beta_1 g_t', \\
n_t = (1 - \beta_3) n_{t-1} + \beta_3 (g_t')^2,
\end{gathered}
\end{equation}
where $\beta_3$ is the third decay rate, and $g_t'$ represents a modified gradient based on gradient difference term $g_t - g_{t-1}$, defined as
\begin{equation}
g_t' = g_t + (1 - \beta_1)(g_t - g_{t-1}).
\label{eq:adan_gprime}
\end{equation}

Then, the update rule of Adan is expressed as
\begin{equation}
w_{t+1} = w_t - \frac{\eta}{\sqrt{n_t + \epsilon}} m_t.
\label{eq:adan_simple}
\end{equation}

In the original paper, this version of Adan is referred to as Vanilla Adan, which defines its fundamental principles and theoretical framework. However, the implementation of the Adan algorithm is presented in a step-by-step manner as follows
\begin{equation}
\begin{gathered}  
m_t = (1 - \beta_1) m_{t-1} + \beta_1 g_t, \\
v_t = (1 - \beta_2) v_{t-1} + \beta_2 (g_t - g_{t-1}), \\
n_t = (1 - \beta_3) n_{t-1} + \beta_3 [g_t + (1 - \beta_2)(g_t - g_{t-1})]^2, \\
\eta_t = \frac{\eta}{\sqrt{n_t + \epsilon}}, \\
w_{t+1} = (1 + \lambda_t \eta)^{-1} [w_t - \eta_t (m_t + (1 - \beta_2) v_t)].
\label{eq:adan}
\end{gathered}
\end{equation}

The algorithm also includes a weight restart condition, which is not discussed here. In the original paper, experiments are conducted on various neural networks (e.g., ResNet, ViT, GPT-2) to evaluate the performance of Adan, demonstrating its consistent effectiveness. The default hyperparameter values used for Adan are set as $\beta_1 = 0.02$, $\beta_2 = 0.08$, $\beta_3 = 0.01$, and $\lambda = 0.02$. These default values can be adjusted for fine-tuning, and the learning rate can be set according to the specific network.

\subsection{Evolved Sign Momentum}
Many optimizers, such as Adam, Nadam, and Adafactor, have been developed through manual design and experimental tuning. Another approach is to define the discovery of deep learning optimizers as a program search. Evolved sign momentum (Lion) is an effective optimizer developed using genetic algorithms \cite{chen2023}. Lion utilizes a variety of methods including evolutionary search with warm-start and restart, pruning through abstract execution, funnel selection, and program simplification. Unlike classical adaptive optimizers such as Adam, it uses the sign function to control the update size and does not require the second moment estimate of gradients; therefore, it is more memory efficient. In Lion, $\beta_2$ is used instead of $\beta_1$ for the update of $m_t$, and an additional momentum variable ($m_{L,t}$) is defined as
\begin{equation}
\begin{gathered}
m_t = \beta_2 m_{t-1} + (1 - \beta_2) g_t, \\
m_{L,t} = \beta_1 m_{t-1} + (1 - \beta_1) g_t.
\end{gathered}
\end{equation}

Similar to AdamW, the Lion update rule applies the weight decay as
\begin{equation}
w_{t+1} = w_t - \eta_t (\text{sign}(m_{L,t}) + \lambda w_t),
\label{eq:lion}
\end{equation}
where $\text{sign}(\cdot)$ is the sign operator, and the weights are updated according to the sign of $m_{L,t}$. This simplified update rule allows Lion to learn faster and more efficiently with lower memory requirements. Lion is successfully used in Google's ads Clickthrough rate model. The discrete nature of the sign function in the original version of Lion can cause convergence problems in some models. To address this issue, the Refined Lion optimizer (RLion) is proposed with an update rule incorporating the arctangent function and the product of a scaling factor and momentum \cite{rong2025}.

\subsection{Second-Order Clipped Stochastic Optimization}
Most optimizers, such as Adam and its variants, use only first-order information, i.e., the gradient of the loss function. In contrast, Second-order clipped stochastic optimization (Sophia) is an optimizer based on second-order information, specifically the Hessian that is the second derivative of the loss function \cite{liu2023}. However, since directly computing the Hessian matrix is computationally expensive, Sophia approximates this information. Sophia's core approach combines techniques such as diagonal Hessian approximation, Fisher information matrix estimation, gradient clipping, and a hybrid mini-batch strategy, making second-order optimization scalable and practical. In this way, Sophia aims to accelerate convergence in transformer-based large language models (LLMs), while reducing both training time and energy consumption. Sophia usually uses decoupled weight decay similar to AdamW in the update. The Sophia update rule can be expressed in two steps, as follows
\begin{equation}
\begin{gathered}
w_t = w_t - \eta_t \lambda w_t, \\
w_{t+1} = w_t - \eta_t \text{clip}(m_t / \max\{\delta h_t, \epsilon\}, \tau_c),
\label{eq:sophia}
\end{gathered}
\end{equation}
where $\text{clip}(\cdot)$ is the clipping function applied element-wise, $\text{clip}(z, \tau_c) = \max\{\min\{z, \tau_c\}, -\tau_c\}$. For any variables $z$ and $\tau_c$, this function ensures that the $z$ value remains between $\tau_c$ and $-\tau_c$. Here, $\tau_c$ denotes the clipping threshold, which can be set to 1 in equation \eqref{eq:sophia}. The parameter $\delta$ is used as the clipping rate. $h_t$ is the exponential moving average of diagonal Hessian estimates. It is updated at every $k$ step as ($k$ is typically set to 10)
\begin{equation}
h_t = \beta_2 h_{t-k} + (1 - \beta_2) \hat{h}_t,
\label{eq:sophia_h}
\end{equation}
where $\hat{h}_t$ represents the diagonal Hessian estimate. There are two options to obtain $\hat{h}_t$: Hutchinson's unbiased estimator and Gauss-Newton-Bartlett (GNB) estimator, $\hat{h}_t = \text{Estimator}(w_t)$. The default hyperparameter values used for Sophia are set as $\beta_1 = 0.96$, $\beta_2 = 0.99$, $\epsilon = 10^{-12}$, $\tau_c=1$, $\delta = 0.01, 0.05$, and $\lambda = 0.02$. As a result, for language modeling with GPT models, Sophia demonstrates up to a 2x improvement over AdamW in the total computational cost and training time.

\subsection{Momentum Orthogonalized by Newton-Schulz}
Deep learning optimization algorithms typically update the weights in all layers of the neural network architecture using the same method. Unlike these, Momentum Orthogonalized by Newton-Schulz (Muon) is an innovative optimizer that takes a hybrid approach to updating weights based on the layer type \cite{jordan2024}. The Muon optimizer uses orthogonalization, particularly for the 2D weights in the hidden layers, while it uses AdamW for other weights, such as scalars, vectors, and output embedding-layer weights. In Muon, the first moment estimate of gradients ($m_t$) is computed as in \eqref{eq:muon}, then a Newton-Schulz iteration is applied to the resulting moment, and finally the weight update is performed using the orthogonalized momentum ($o_t$) as
\begin{equation}
\begin{gathered}
m_t = \gamma m_{t-1} + g_t, \\
o_t = \text{NewtonSchulz}(m_t), \\
w_{t+1} = w_t - \eta o_t.
\label{eq:muon}
\end{gathered}
\end{equation}

Here, $\text{NewtonSchulz}(\cdot)$ is an operator that performs the Newton-Schulz iteration. It enables the approximate orthogonalization of the moment, i.e.\ the update matrix. The purpose of orthogonalizing the update matrix is to ensure that learning occurs not only in the dominant few directions but also in the rare directions. This makes the learning dynamics more stable, allowing models to learn faster. The empirical results show that training with Muon requires less time than with AdamW on a model containing 1.5 billion parameters. Moreover, Muon demonstrates strong scalability when applied to the training of LLMs \cite{liu2025}.

\subsection{Neuron-wise Normalized Muon}
Muon improves the conditioning of update matrices through orthogonalization. However, the resulting update norms may still differ substantially across neurons. Neuron-wise Normalized Muon (NorMuon) extends Muon by incorporating neuron-level adaptive learning rates after the orthogonalization step \cite{li2025normuon}. For a two-dimensional hidden-layer weight matrix $w_t \in \mathbb{R}^{m \times n}$, its update rule can be summarized as
{
\begin{equation}
\begin{gathered}
m_t = \beta_1 m_{t-1} + (1-\beta_1)g_t, \\
o_t = \text{NewtonSchulz}(m_t), \\
v_t = \beta_2 v_{t-1} + (1-\beta_2)
\text{mean}_{\text{cols}}(o_t \odot o_t), \\
V_t = \text{ExpandRows}(v_t), \\
\hat{o}_t = o_t \oslash (\sqrt{V_t}+\epsilon), \\
\hat{\eta}_t = 0.2\eta\frac{\sqrt{mn}}{\|\hat{o}_t\|_F}, \\
w_{t+1} = w_t - \eta\lambda w_t - \hat{\eta}_t\hat{o}_t.
\label{eq:normuon}
\end{gathered}
\end{equation}
}
\par\indent Here, $\odot$ and $\oslash$ denote element-wise multiplication and division, respectively. The operator $\text{mean}_{\text{cols}}(\cdot)$ computes a row-wise mean over the columns, while $\text{ExpandRows}(\cdot)$ broadcasts the resulting vector to match the size of the weight matrix. Thus, $v_t \in \mathbb{R}^{m}$ stores a second-moment statistic for each neuron with a limited memory overhead. The Frobenius norm $\|\cdot\|_F$ is used to rescale the normalized update. The authors also propose an implementation compatible with the Fully Sharded Data Parallel framework. In experiments on the pretraining of a model with 1.1 billion parameters, the authors report 21.74\% better training efficiency than Adam and an 11.31\% improvement over Muon.

\subsection{\ScheduleFreePlus}
A learning-rate schedule is commonly used to reduce the learning rate as training progresses. \ScheduleFreePlus{} is a recent optimizer proposed for LLM training that replaces a manually selected decay schedule with iterate averaging and a Polyak-like adaptive step size \cite{defazio2026schedulefreeplus}. It extends Schedule-Free Learning \cite{defazio2024schedulefree} to larger batch sizes and model sizes by reintroducing inner momentum, warm-starting the averaging process, and adjusting the interpolation parameter during long training runs. \ScheduleFreePlus{} maintains a raw iterate $z_t$, an averaged iterate $x_t$, and an interpolation point $y_t$. Using the notation of this paper, the effective step size $\alpha_t$ can be expressed as

\begin{equation}
\alpha_t =
\eta_t^{\text{wup}}
\frac{
\max\left\{0,\,
L(y_t)-L^\star+
\beta_t\left\langle \nabla_w L(y_t), z_t-x_t \right\rangle
\right\}
}{
\sqrt{\pi/2}\,
\widehat{\left\|\nabla_w L(y_t)\right\|_1}
},
\label{eq:schedulefreeplus_step}
\end{equation}

\par\noindent where $L^\star$ is an estimate of the optimal loss value, $\eta_t^{\text{wup}}$ denotes the warm-up multiplier, and the denominator uses an exponential moving average of the $L_1$ norm of the gradient. The interpolation and update steps are stated as
{
\begin{equation}
\begin{gathered}
y_t = (1-\beta_t)z_t + \beta_t x_t, \\
m_t = \beta_1 m_{t-1} + (1-\beta_1)g_t, \\
v_t = \beta_2 v_{t-1} + (1-\beta_2)g_t^2, \\
z_{t+1} =
z_t - \alpha_t^2\lambda y_t
- \alpha_t\frac{\hat{m}_t}{\sqrt{\hat{v}_t+\epsilon}}, \\
x_{t+1} = (1-c_{t+1})x_t + c_{t+1}z_{t+1}.
\label{eq:schedulefreeplus}
\end{gathered}
\end{equation}
}
\par\indent Here, $c_{t+1}$ is the averaging coefficient, and $\beta_t$ controls the interpolation between the raw and averaged iterates. The squared dependence of the weight-decay term on $\alpha_t$ follows the fully decoupled AdamC formulation used in \ScheduleFreePlus. The final averaged iterate is returned as the model estimate. In experiments with 1,000 training tokens per parameter, the author reports that \ScheduleFreePlus{} reaches the same loss as a linear-decay schedule with a 31\% reduction in training time. The reported final-loss advantage is associated with long training runs and is not observed for shorter runs of 20--100 tokens per parameter.

\section{Illustrative Trajectory Analysis of Optimizer Behavior}

This section uses trajectory plots to illustrate how optimizer mechanisms can shape the path followed on a controlled non-convex surface. The purpose of these visualizations is to provide intuitive insights: they help connect update rules with observable behaviors such as smoothness, oscillation, overshooting, damping, and rapid correction. They should not be interpreted as a general empirical ranking of optimizers for deep neural network training, where architecture, data, batch size, regularization, schedule design, and hyperparameter tuning strongly affect performance.

\subsection{Visualization Protocol}

The illustrative benchmark used in this section is the Beale function,
\begin{equation}
B(x,y)= (1.5-x+xy)^2 +(2.25-x+xy^2)^2 +(2.625-x+xy^3)^2,
\label{eq:beale}
\end{equation}

\par\noindent which has a global minimum at $(x,y)=(3,0.5)$ with $B(x,y)=0$. The function is useful for visualization because its narrow valleys and steep regions make changes in step direction and step size visible in two dimensions. The optimizers are grouped by mechanism, and the starting points are selected to make the characteristic trajectory differences within each group visible. Therefore, the iteration counts reported below should be read within each figure-specific setting, not as a universal comparison across all optimizers.

\subsection{Optimizer Trajectory Comparisons}

\begin{figure}[htbp]
    \centering    \includegraphics[width=0.48\textwidth]{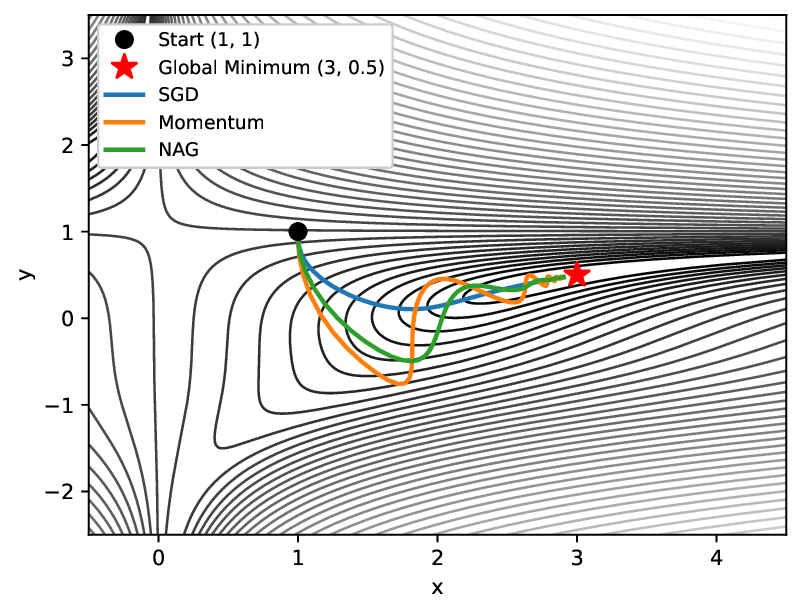}
    \caption{Illustrative trajectories of SGD, Momentum, and NAG optimizers on the Beale function, starting from $(1,1)$ toward the global minimum at $(3,0.5)$}    \label{fig:optimization_paths1}
\end{figure}

\begin{figure}[htbp]
    \centering    \includegraphics[width=0.48\textwidth]{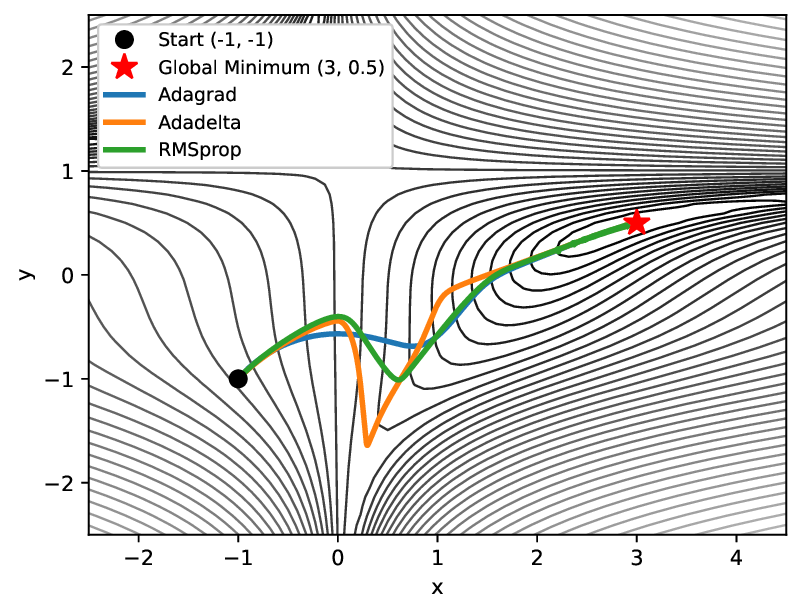}
    \caption{Illustrative trajectories of Adagrad, Adadelta, and RMSprop optimizers on the Beale function}    \label{fig:optimization_paths2}
\end{figure}

Fig.~\ref{fig:optimization_paths1} shows the trajectories of SGD, Momentum, and NAG as fundamental gradient-based optimizers on the Beale function, starting from $(1,1)$ toward the global minimum at $(3,0.5)$. In this illustrative setting, SGD follows a smooth but slow path and requires 921 update time steps to converge. This behavior is consistent with the absence of an accumulated-gradient mechanism. Momentum accelerates convergence by using past gradients and reaches the minimum in 77 update time steps, although oscillation appears near curved regions of the loss surface. NAG further stabilizes the momentum behavior through forward gradient evaluation and reaches the minimum in 61 update time steps. The visualization highlights how momentum and lookahead mechanisms can alter both convergence speed and path shape on a sharply curved surface. Fig.~\ref{fig:optimization_paths2} illustrates Adagrad, Adadelta, and RMSprop as baseline adaptive optimizers, starting from $(-1,-1)$. Adagrad follows a smooth but very slow trajectory, requiring 8,719 update time steps, which reflects the effect of its continuously accumulated squared-gradient term. Adadelta reduces this slowdown by adapting step sizes and converges in 5,488 update time steps, although deviations are visible in high-slope regions. RMSprop reaches the minimum in 450 update time steps and maintains a more direct trajectory in this example. The comparison visually separates cumulative adaptive scaling from exponentially decayed adaptive scaling. 

\begin{figure}[htbp]
    \centering    \includegraphics[width=0.48\textwidth]{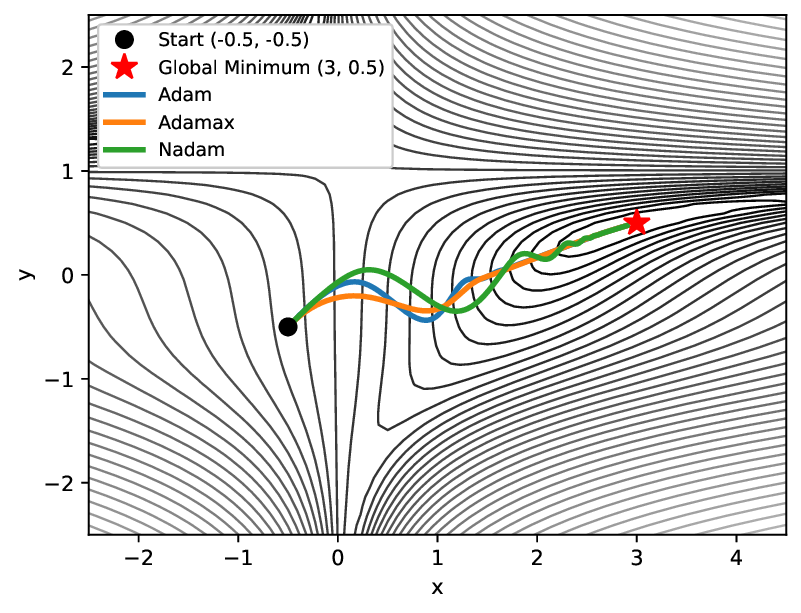}
    \caption{Illustrative trajectories of Adam, Adamax, and Nadam optimizers on the Beale function}    \label{fig:optimization_paths3}
\end{figure}

\begin{figure}[htbp]
    \centering    \includegraphics[width=0.48\textwidth]{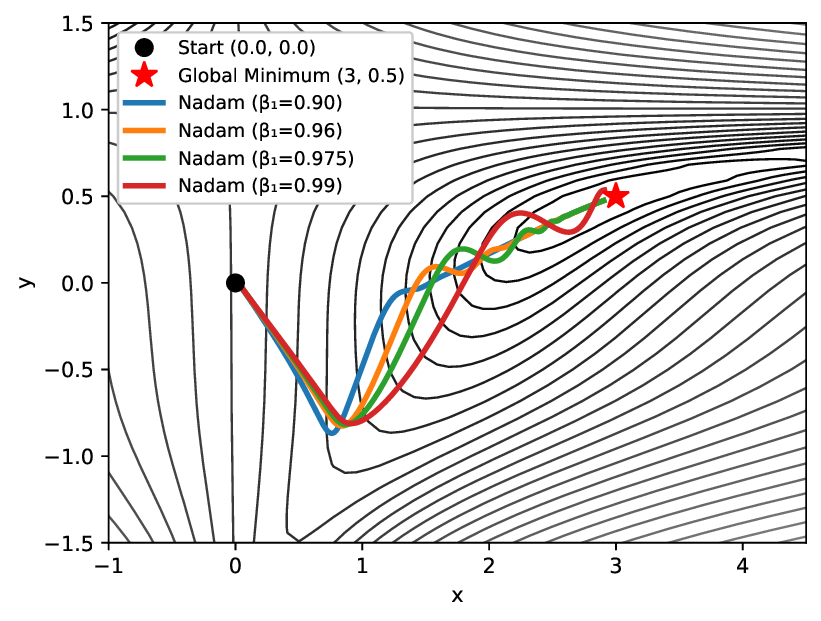}
    \caption{Illustrative Nadam trajectories for different values of the first decay rate, $\beta_1=0.90, 0.96, 0.975, 0.990$}    \label{fig:optimization_paths4}
\end{figure}

Fig.~\ref{fig:optimization_paths3} presents Adam, Adamax, and Nadam from $(-0.5,-0.5)$. All three adaptive methods reach the minimum in this setting, but they do so with different path shapes and iteration counts. Adam reaches the minimum in 1,027 update time steps by combining adaptive learning rates with first moment information. Adamax follows a smoother and more conservative path, requiring 1,603 update time steps. Nadam converges in 848 update time steps, suggesting that the Nesterov-style correction can provide a more proactive update direction on this landscape. This observation is specific to the chosen surface and settings. Fig.~\ref{fig:optimization_paths4} illustrates the effect of the first decay rate hyperparameter $\beta_1$ on Nadam from $(0,0)$. Lower values such as $\beta_1=0.90$ produce a more damped response and require 1,058 update time steps. As $\beta_1$ increases to 0.96 and 0.975, the trajectories become more oscillatory, while the required iterations decrease to 980 and 895, respectively. The most aggressive setting, $\beta_1=0.99$, reaches the target in 312 update time steps but with substantial overshooting. The figure therefore visualizes the familiar trade-off between damping and acceleration in momentum-like mechanisms.

\begin{figure}[htbp]
    \centering    \includegraphics[width=0.48\textwidth]{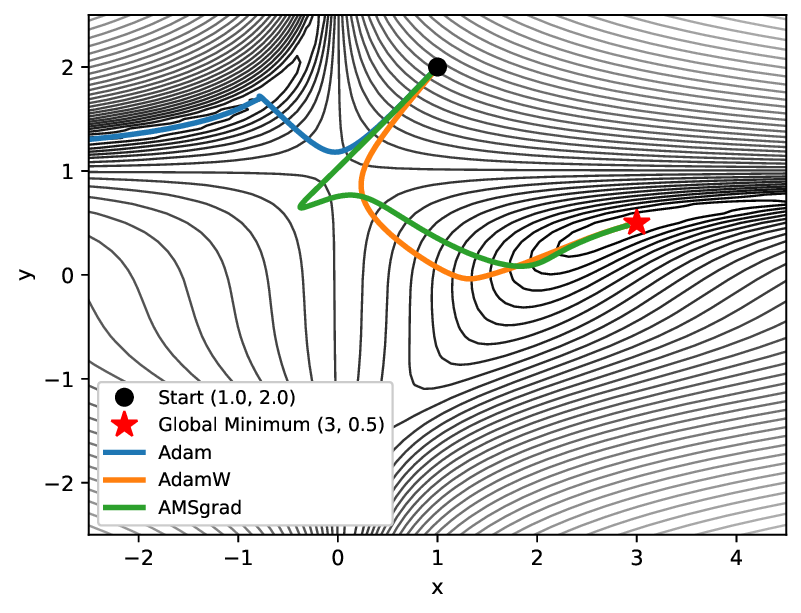}
    \caption{Illustrative trajectories of Adam, AdamW, and AMSgrad optimizers, showing a setting where Adam does not reach the target while its variants do}    \label{fig:optimization_paths5}
\end{figure}

\begin{figure}[htbp]
    \centering    \includegraphics[width=0.48\textwidth]{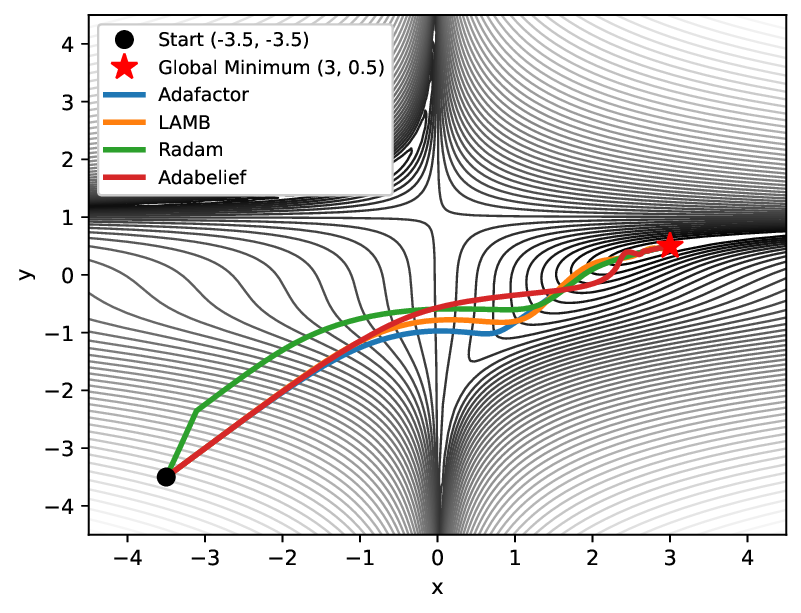}
    \caption{Illustrative trajectories of Adafactor, LAMB, Radam, and Adabelief optimizers on the Beale function}    \label{fig:optimization_paths6}
\end{figure}

Fig.~\ref{fig:optimization_paths5} compares Adam, AdamW, and AMSgrad in a setting where the standard Adam trajectory does not reach the target region. Starting from $(1,2)$, Adam becomes trapped away from the global minimum under the selected settings. AdamW reaches the minimum in 14,336 update time steps, while AMSgrad reaches it in 2,157 update time steps. The visualization suggests how decoupled weight decay and the use of a maximum of past squared gradients can change stability on this surface. It does not by itself establish general superiority for neural network training. Fig.~\ref{fig:optimization_paths6} illustrates Adafactor, LAMB, Radam, and Adabelief as advanced Adam-related optimizers from $(-3.5,-3.5)$. In this setting, all four methods converge, but their paths differ. Adabelief reaches the minimum in 250 update time steps by adapting its step size according to the discrepancy between predicted and observed gradients. LAMB follows closely with 257 update time steps and exhibits a direct layerwise-normalized path. Adafactor converges in 360 update time steps and remains comparatively conservative. Radam requires 1,340 update time steps and takes a wider detour before convergence. The figure is useful for contrasting belief-based, layerwise, factored-memory, and variance-rectified adaptive mechanisms.

\begin{figure}[htbp]
    \centering    \includegraphics[width=0.48\textwidth]{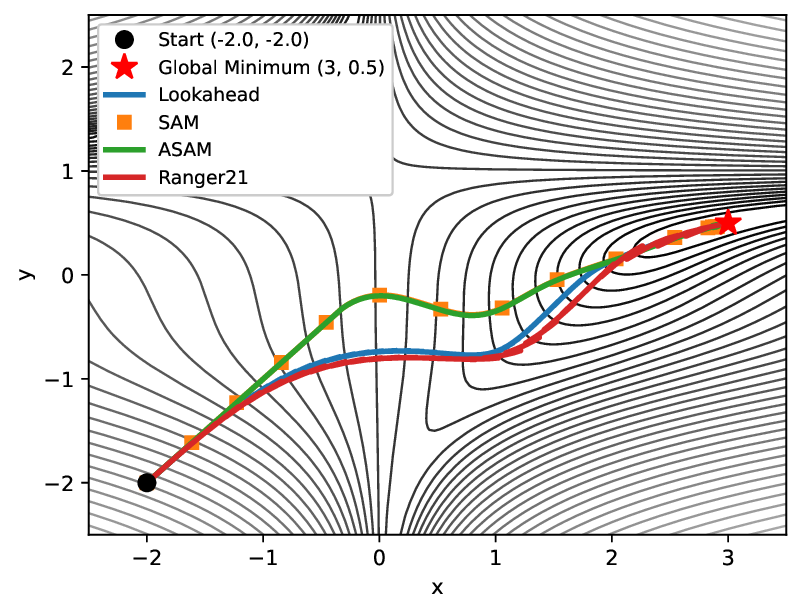}
    \caption{Illustrative trajectories of Lookahead, SAM, ASAM, and Ranger21 optimizers on the Beale function}    \label{fig:optimization_paths7}
\end{figure}

\begin{figure}[htbp]
    \centering    \includegraphics[width=0.48\textwidth]{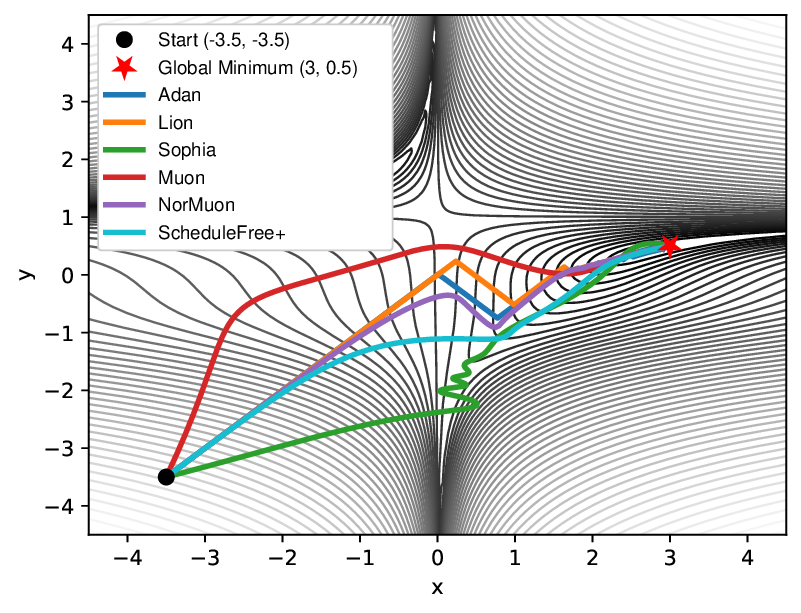}
    \caption{Illustrative trajectories of Adan, Lion, Sophia, Muon, NorMuon, and \ScheduleFreePlus{} optimizers on the Beale function}    \label{fig:optimization_paths8}
\end{figure}

\begin{table*}[htbp]
\renewcommand{\arraystretch}{1.2}
\caption{Summary of the illustrative optimizer behaviors observed on the Beale-function visualizations}
\begin{center}
\begin{tabular}{|>{\raggedright\arraybackslash}p{3.3cm}|>{\raggedright\arraybackslash}p{3.8cm}|>{\raggedright\arraybackslash}p{3.8cm}|>{\raggedright\arraybackslash}p{3.8cm}|}
\hline
\textbf{Optimizer Group}&\textbf{Main Mechanism}&\textbf{Visual Behavior in This Section}&\textbf{Key Insight} \\
\hline
SGD, Momentum, NAG & Gradient descent, accumulated gradients, lookahead & Momentum-based methods reduce the number of steps but may introduce oscillation & Acceleration and damping must be balanced \\ \hline
Adagrad, Adadelta, RMSprop & Adaptive scaling by squared gradients & Cumulative scaling is conservative, while decayed scaling produces shorter paths & The memory of past gradients controls step shrinkage \\ \hline
Adam, Adamax, Nadam & First and second moment adaptation & Nesterov-style correction changes the path and can reduce steps in this setting & Adaptive momentum variants differ in direction and scaling \\ \hline
Nadam hyperparameter variants & First moment decay tuning & Higher $\beta_1$ accelerates but increases overshooting & Hyperparameters affect both speed and stability \\ \hline
Adam, AdamW, AMSgrad & Decoupled weight decay and long-term second moment control & Variants recover convergence where Adam stalls under the selected settings & Stabilizing modifications can change failure modes \\ \hline
Adafactor, LAMB, Radam, Adabelief & Memory factorization, layerwise scaling, variance rectification, belief correction & Paths differ in directness and conservativeness & Similar Adam-family optimizers can behave visibly differently \\ \hline
Lookahead, SAM, ASAM, Ranger21 & Slow-fast weights, sharpness-aware updates, hybrid scheduling & Search behavior becomes more structured and sometimes more direct & Composite mechanisms can reshape trajectories substantially \\ \hline
Adan, Lion, Sophia, Muon, NorMuon, \ScheduleFreePlus & Nesterov estimation, sign updates, curvature approximation, orthogonalization, neuron-wise normalization, schedule-free averaging & Sign-based, curvature-based, orthogonalized, normalized, and averaged updates produce distinct path geometries & Recent optimizers can be introduced through update geometry and step-size control \\
\hline
\end{tabular}
\label{tab:behavior_summary}
\end{center}
\end{table*}

Fig.~\ref{fig:optimization_paths7} shows Lookahead, SAM, ASAM, and Ranger21 as optimizers or optimizer frameworks that modify the local search behavior beyond a standard adaptive update. Starting from $(-2,-2)$, ASAM reaches the minimum in 344 update time steps and SAM in 369 update time steps. Their relatively direct paths illustrate the effect of sharpness-aware perturbations in this landscape. Ranger21 reaches the target in 180 update time steps under the selected settings, reflecting the combined effect of several mechanisms, including adaptive gradient clipping, learning-rate scheduling, and Lookahead. Standalone Lookahead is more conservative and requires 1,246 update time steps. Fig.~\ref{fig:optimization_paths8} presents contemporary optimizers, Adan, Lion, Sophia, Muon, NorMuon, and \ScheduleFreePlus{}, from $(-3.5,-3.5)$. Sophia reaches the minimum in 336 update time steps in this visualization, consistent with the role of curvature information in shaping more corrective steps. Adan reaches the minimum in 593 update time steps and follows a direct path based on Nesterov momentum estimation. Muon and Lion require 1,985 and 2,130 update time steps, respectively, under these settings. Lion's piecewise path visually reflects the sign-based update direction, whereas Muon's smoother curve reflects the orthogonalized momentum update. The NorMuon and \ScheduleFreePlus{} paths are included as two-dimensional illustrative projections: the former highlights neuron-wise normalization following orthogonalization and reaches the target region in 1,516 update time steps, while the latter illustrates an averaged progression associated with schedule-free adaptive step-size control and reaches the target region in 2,099 update time steps. These visual differences help explain mechanisms, but the trajectories should not be generalized as a broad ranking of modern optimizers.

Overall, the trajectory figures provide a visual bridge between the update rules reviewed earlier and the practical behaviors often discussed in optimizer design. They are best read as an explanatory atlas of mechanisms rather than as a substitute for task-level benchmarks on neural network architectures. The following transformer experiments are therefore included as a compact complement that connects selected optimizer mechanisms with validation-loss and throughput behavior in a neural language-modeling setting. To synthesize these figure-specific observations, Table~\ref{tab:behavior_summary} summarizes the main mechanisms, visual behaviors, and key insights for each optimizer group.

\subsection{Small-Scale Transformer Experiments}

To complement the two-dimensional trajectory analysis with limited task-level examples, two small-scale transformer experiments were conducted. A compact character-level decoder-only transformer was trained on the raw WikiText-2 corpus \cite{merity2016pointer}. The model used four transformer blocks, four attention heads, an embedding dimension of 192, a context length of 128, and approximately 1.90 million trainable parameters. These experiments are not intended as comprehensive benchmarks; rather, they provide controlled auxiliary checks for optimizer behavior on a small transformer language-modeling task.

\paragraph{Adaptive moment baselines} Adam, AdamW, Nadam, Radam, and Adabelief were evaluated under the same training budget of 3,000 update steps, three random seeds, batch size 64, gradient clipping at 1.0, learning rate $3 \times 10^{-4}$, and weight decay 0.01.

\begin{figure*}[htbp]
    \centering    \includegraphics[width=0.6\textwidth]{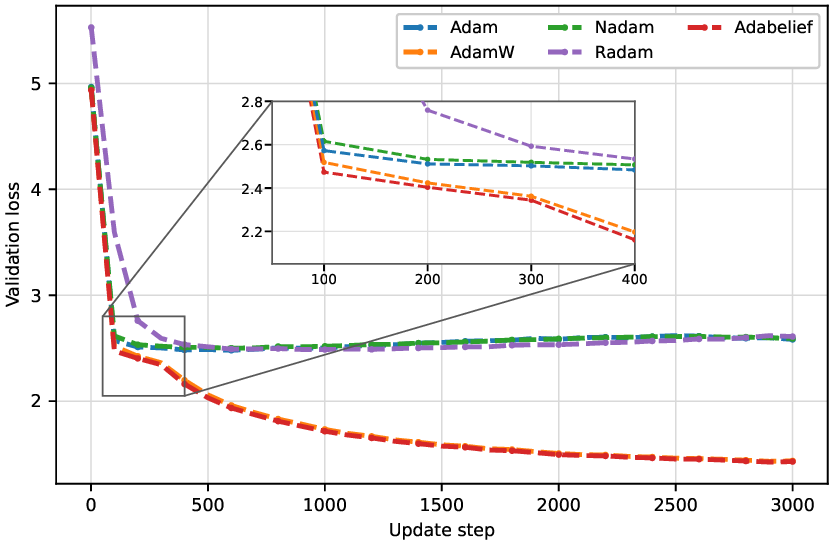}
    \caption{Small-scale transformer experiment for adaptive moment baselines. The dashed curves with small markers show the mean validation loss over three random seeds under the matched 3,000-step training budget. The inset magnifies the early 50--400 update-step region where several curves are closely spaced.} \label{fig:group1_transformer}
\end{figure*}

\begin{table*}[htbp]
\renewcommand{\arraystretch}{1.2}
\caption{Adaptive moment baseline transformer results over three random seeds}
\begin{center}
\begin{tabular}{|>{\centering\arraybackslash}p{2cm}|>{\centering\arraybackslash}p{3cm}|>{\centering\arraybackslash}p{3cm}|>{\centering\arraybackslash}p{2cm}|>{\centering\arraybackslash}p{2cm}|}
\hline
\textbf{Optimizer}&\textbf{Final Valid. Loss}&\textbf{Best Valid. Loss}&\textbf{Best Step}&\textbf{Tokens/s} \\
\hline
Adam & 2.581 $\pm$ 0.013 & 2.480 $\pm$ 0.004 & 533 & 387397 \\ \hline
AdamW & 1.437 $\pm$ 0.005 & 1.430 $\pm$ 0.004 & 2933 & 439791 \\ \hline
Nadam & 2.590 $\pm$ 0.002 & 2.498 $\pm$ 0.006 & 633 & 435499 \\ \hline
Radam & 2.608 $\pm$ 0.014 & 2.484 $\pm$ 0.003 & 967 & 426815 \\ \hline
Adabelief & 1.431 $\pm$ 0.005 & 1.423 $\pm$ 0.004 & 2933 & 334580 \\
\hline
\end{tabular}
\label{tab:group1_transformer_summary}
\end{center}
\end{table*}

As shown in Fig.~\ref{fig:group1_transformer} and Table~\ref{tab:group1_transformer_summary}, AdamW and Adabelief continue to reduce validation loss across the 3,000-step run, whereas Adam, Nadam, and Radam plateau at higher validation losses under the same nominal hyperparameter setting. Adabelief obtains the lowest mean final validation loss, $1.431 \pm 0.005$, followed closely by AdamW with $1.437 \pm 0.005$. In contrast, Adam, Nadam, and Radam remain near 2.6 final validation loss. The throughput values indicate the measured training speed in tokens per second: AdamW and Nadam are the fastest in this group, while Adabelief is slower, consistent with the additional belief-based second-moment computation. The inclusion of Nadam therefore provides a useful comparison for Nesterov-style adaptive momentum, but in this compact configuration it does not close the gap to the decoupled-weight-decay and belief-based variants. These results should be interpreted narrowly: they support the qualitative point that closely related adaptive moment optimizers can differ substantially even in a small transformer setting, but they do not replace larger-scale benchmarks or optimizer-specific hyperparameter tuning.

\paragraph{Contemporary optimizers} The second experiment considered the contemporary optimizer group using the same corpus, architecture, random seeds, batch size, evaluation protocol, gradient clipping, and weight-decay setting as in the adaptive moment baseline experiment, under the same 3,000-update-step budget to better expose late-stage behavior. The evaluated optimizers were Adan, Lion, Sophia, Muon, NorMuon, and \ScheduleFreePlus. For this controlled task-level comparison, the Muon and NorMuon implementations apply Newton-Schulz orthogonalization to two-dimensional hidden-layer weight matrices, while the remaining parameters are updated with AdamW-style adaptive updates. NorMuon additionally applies row-wise adaptive normalization after orthogonalization. The Sophia implementation uses diagonal curvature clipping with a gradient-square proxy, and \ScheduleFreePlus{} is represented by a schedule-free adaptive update with warm-up and iterate averaging. These implementations are intended as small-scale mechanism-level projections of the reviewed methods rather than full reproductions of the large-scale training systems for which several of these optimizers were originally proposed.

\begin{figure*}[htbp]
    \centering    \includegraphics[width=0.6\textwidth]{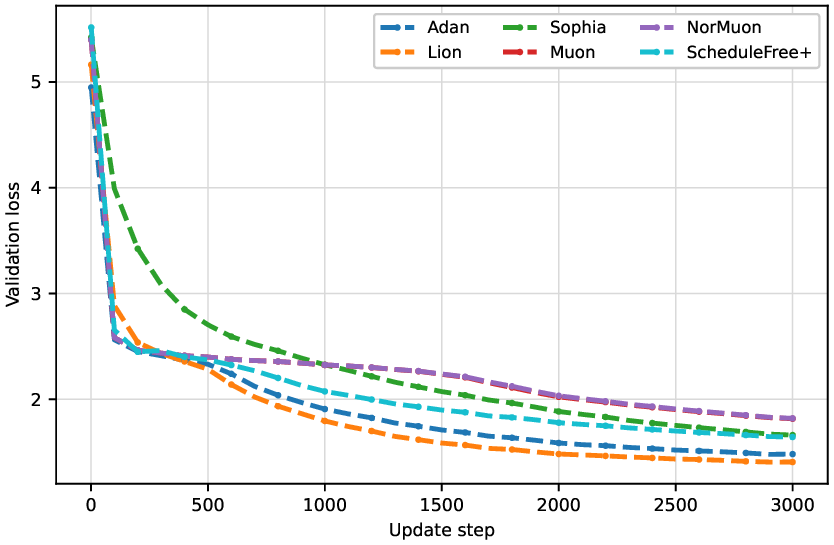}
    \caption{Small-scale transformer experiment for contemporary optimizers. The dashed curves with small markers show the mean validation loss over three random seeds under the matched training budget.}   \label{fig:group2_transformer}
\end{figure*}

\begin{table*}[htbp]
\renewcommand{\arraystretch}{1.2}
\caption{Contemporary optimizer transformer results over three random seeds}
\begin{center}
\begin{tabular}{|>{\centering\arraybackslash}p{2cm}|>{\centering\arraybackslash}p{3cm}|>{\centering\arraybackslash}p{3cm}|>{\centering\arraybackslash}p{2cm}|>{\centering\arraybackslash}p{2cm}|}
\hline
\textbf{Optimizer}&\textbf{Final Valid. Loss}&\textbf{Best Valid. Loss}&\textbf{Best Step}&\textbf{Tokens/s} \\
\hline
Adan & 1.480 $\pm$ 0.003 & 1.476 $\pm$ 0.006 & 2933 & 269599 \\ \hline
Lion & 1.407 $\pm$ 0.004 & 1.402 $\pm$ 0.005 & 2967 & 352776 \\ \hline
Sophia & 1.662 $\pm$ 0.010 & 1.661 $\pm$ 0.009 & 3000 & 315436 \\ \hline
Muon & 1.816 $\pm$ 0.005 & 1.815 $\pm$ 0.005 & 3000 & 205421 \\ \hline
NorMuon & 1.819 $\pm$ 0.005 & 1.819 $\pm$ 0.005 & 3000 & 181964 \\ \hline
\ScheduleFreePlus{} & 1.644 $\pm$ 0.008 & 1.642 $\pm$ 0.007 & 2967 & 289990 \\
\hline
\end{tabular}
\label{tab:group2_transformer_summary}
\end{center}
\end{table*}

As shown in Fig.~\ref{fig:group2_transformer} and Table~\ref{tab:group2_transformer_summary}, Lion obtains the lowest mean final validation loss in this controlled setting, $1.407 \pm 0.004$, followed by Adan with $1.480 \pm 0.003$. \ScheduleFreePlus{} and Sophia reach $1.644 \pm 0.008$ and $1.662 \pm 0.010$, respectively, while Muon and NorMuon reach similar higher values of $1.816 \pm 0.005$ and $1.819 \pm 0.005$. Under the 3,000-step budget, the best step moves away from the final update for Lion, Adan, and \ScheduleFreePlus{}, whereas Sophia, Muon, and NorMuon still obtain their best validation loss at step 3,000. The measured Tokens/s values also highlight the computational trade-off among contemporary methods: Lion and Sophia provide comparatively high training throughput, whereas Muon and NorMuon are slower because the orthogonalization and normalization steps add update-time overhead. The results should therefore be read as a compact task-level illustration: under this small transformer configuration, sign-based and Nesterov-difference-based updates are comparatively effective, whereas matrix-orthogonalization mechanisms may require larger models, longer runs, or more optimizer-specific tuning to show their intended advantages.

\section{Discussion}
\subsection{Evolution of Optimizers}
The development of optimizers should be examined in parallel with the historical development of deep learning. Following periods of stagnation, often referred to as AI winters, deep learning experienced a significant revival in the 2010s and gained considerable momentum with the emergence of LLMs. Within this context, the evolution of deep learning optimizers shows a progression from classical methods to adaptive approaches, and then to scalable methods. Based on the optimizers detailed in the previous section, this progression can be analyzed across several phases and key milestones.

Cauchy's work must first be emphasized as the foundational contribution that initiated the concept of gradient descent \cite{cauchy1847,lemarechal2012}. Gradient descent was adapted into deep learning through SGD to form a foundational framework for training artificial neural networks. The SGD optimizer enables the iterative update of parameters or weights in neural networks by defining the direction and size of each optimization step \cite{ruder2016,robbins1951,bottou2010}. Although SGD is simple and stable, its high oscillation and slow convergence during optimization are considered major drawbacks. To address these issues, the concept of 'momentum' from physics was introduced, incorporating the influence of past gradients into the update process. SGD with momentum or simply Momentum aims to reduce the risk of getting trapped in local minima by determining the update step based on both current and past gradients \cite{rumelhart1986, polyak1964,qian1999}. On the other hand, NAG improves Momentum by introducing a lookahead mechanism, i.e., it anticipates the next position and computes the gradient at that point before performing the weight update. This approach enables more accurate gradient estimation and facilitates a more controlled convergence during optimization \cite{nesterov1983,sutskever2013}. In the most basic optimizers --- SGD, Momentum, and NAG --- a single learning rate had to be manually set for all weights. This posed challenges when dealing with data containing both sparse and dense features.

With the introduction of Adagrad in 2011, adaptive learning rates began to be used in deep learning. Adagrad enables each weight to be updated with an individual learning rate, scaled by the accumulation of squared gradients \cite{duchi2011}. This allows all weights to be optimized in a balanced manner and improves efficiency in processing sparse data. However, as the learning rate continuously decreases over time and approaches zero, the learning process may eventually stall. Adadelta and RMSprop addressed the diminishing learning rate problem of AdaGrad. While RMSprop uses an exponentially decaying average of squared gradients, Adadelta eliminates the dependence on the learning rate hyperparameter \cite{hinton2012,zeiler2012}. Introduced in 2014, Adam represented a significant milestone in the development of optimization algorithms and became the default optimizer in many deep learning applications. Adam combines the concept of momentum from Momentum (first moment) with the concept of averaging squared gradients from RMSprop (second moment) \cite{kingma2014}. As a result, faster convergence was achieved with less manual hyperparameter tuning, and heterogeneous data could be processed more efficiently. However, although training with Adam is fast, it may sometimes lead to overfitting, posing a risk to generalization performance.

Although Adam is highly popular, numerous variants, such as Nadam, AdamW, AMSgrad, and Radam, have been developed to overcome its limitations. AdamW, introduced in 2017, has a significant role among these optimizers. It applies weight decay as a separate update component, resulting in improved generalization performance and becoming the standard optimizer for many prominent LLMs \cite{loshchilov2019}. Following AdamW, optimizers such as LARS/LAMB, Lookahead, Adabelief, and SAM/ASAM have contributed to model training by introducing the concepts of 'layerwise adaptive', 'slow/fast weights', 'belief', and 'sharpness', respectively \cite{you2017,you2020,zhang2019lookahead,zhuang2020,foret2021,kwon2021}. In addition, various techniques have been developed to close the generalization gap of adaptive optimizers and to improve training efficiency. Ranger21, aiming for more robust optimization, builds on AdamW as its core optimizer and integrates the following techniques: Adaptive gradient clipping, Gradient centralization, Positive-negative momentum, Linear learning rate warm-up, Explore-exploit learning rate schedule, Stable weight decay, Norm loss, and Lookahead \cite{wright2021}.

In recent years, with the emergence of large-scale models, the primary focus in optimization has shifted to computational cost and memory efficiency. Optimizers such as Adan, Lion, Sophia, and Muon aim to train these models, often with billions of parameters, using minimal resources \cite{xie2024,chen2023,liu2023,jordan2024, rong2025, liu2025}. Adan reduces the number of training epochs by incorporating Nesterov momentum estimation without introducing additional computational overhead, while Lion significantly reduces memory consumption by computing only the sign of the momentum. Sophia utilizes a low-cost approximation of second-order information (the Hessian matrix), promising faster convergence than Adam, particularly in large language models. Muon, on the other hand, has recently attracted attention due to its orthogonalization approach and scalability to LLMs.
Recent developments further extend this progression in two different directions. NorMuon combines Muon's orthogonalization with neuron-level adaptive learning rates to balance update norms across neurons \cite{li2025normuon}. \ScheduleFreePlus{} focuses on reliable LLM training with less tuning by combining iterate averaging with a Polyak-like adaptive step size and eliminating the need for a manually designed learning-rate decay schedule \cite{defazio2026schedulefreeplus}.
In addition to all these, numerous studies have been proposed over the past four years focusing on optimization algorithms \cite{defazio2023,reyad2023,ivgi2023,mishchenko2023,yang2023,zhao2024, zhang2024adam, kristiansen2024narrowing, liang2024cautious, ahn2025dion, si2025adamuon, liu2025cosmos, vyas2024soap, zhu2024apollo}. With continuous improvements, each new study aims to address the shortcomings of previous optimizers and meet emerging requirements.

\subsection{Open Challenges and Future Directions}
The evolution of deep learning optimizers has been primarily directed toward achieving better generalization, more stable and faster convergence, reduced dependence on hyperparameters, and lower resource consumption. A model obtained through optimization with training data must be capable of adapting to previously unseen real-world data, that is, it must possess generalization ability. Achieving a balance between training and test performance represents one of the most critical challenges in the field of optimization. Training is expected to progress toward the minimum in a stable and smooth manner, while also being achieved within fewer iterations and less time. Stability often requires more conservative updates, which, however, may hinder training speed. Ensuring both stable and fast convergence constitutes a significant challenge in reaching a satisfactory solution. Optimizers employ a variety of hyperparameters, such as learning rate, decay rates, and smoothing terms. Although default values of these hyperparameters are commonly used in optimization, they do not always yield satisfactory performance. Achieving optimal performance requires careful, problem-specific tuning of these hyperparameters, which constitutes a considerable challenge in practice. For large-scale models, the memory consumption and computational cost of optimizers are of critical importance. While classical methods often prove inadequate for such models, advanced optimizers may impose high resource requirements. Such a condition can restrict the applicability of novel methods in large-scale deep learning. Consequently, the development of more efficient and scalable optimizers specifically tailored for large-scale models emerges as another key challenge.

Future research can be expected to be directed by the aforementioned challenges in parallel with the advancement of deep learning models. In this context, the objective of future studies will be to enhance generalization ability through stable and fast training processes, while achieving minimal use of hyperparameters and resources. Along this path, a variety of methods or approaches can be adopted. These approaches may focus on storing, computing, and estimating gradients or other information used in optimization more efficiently. In addition, new techniques can be developed for the utilization of second-order information. Moreover, research can also be directed toward meta-learning and learnable optimization algorithms. Improvements can be made in the areas of large batch sizes and distributed training, as well as the design of optimizers that integrate both. To reduce user burden, studies can investigate the automatic adjustment of hyperparameters such as learning rates or the implementation of hyperparameter-free techniques. Also, optimizer designs can be developed to minimize memory, processor, power, and energy demand. Optimizers capable of performing efficiently under resource constraints for embedded and edge devices can also be explored. Research can focus on scaling optimization algorithms for large-scale models. Beyond these methods, quantum-based computation may be employed for training very large models. Furthermore, custom optimizers tailored to specific neural network architectures can be devised.

\section{Conclusion}
In this paper, the methods employed in widely used deep learning optimizers from past to present are investigated. Optimizers have evolved from basic gradient descent methods to today's more sophisticated approaches. Beyond gradient descent, momentum, and adaptive learning rates, numerous techniques have been introduced to facilitate convergence, enhance generalization, and reduce resource consumption. In addition to gradients, studies aimed at obtaining second-order information at low cost have gained importance. The illustrative trajectory analysis and small-scale transformer experiments complement the review by linking optimizer update mechanisms to observable path geometries, validation-loss trends, and training throughput. The evolution of optimizers demonstrates the necessity of developing new methods to adapt to the increasing complexity of deep learning models and the growth of data volumes. It becomes essential to enable the training of models with billions of parameters not only with high generalization ability but also on limited hardware. To this end, future research will continue to focus on multi-objective and resource-efficient optimization tools that guarantee safe convergence while taking model architecture and scale into account.

\bibliographystyle{IEEEtran}
\bibliography{IEEE-bibliography} 

\vspace{0.2cm} 
\hrule
\section*{How to Cite This Work}
\noindent{\footnotesize If you find this review useful, please cite the conference paper:}

\vspace{0.13cm}
\begin{flushleft}
{\footnotesize
@INPROCEEDINGS\{11728119,\\
~~author=\{Altinel, Dogay\},\\
~~booktitle=\{2026 10th International Artificial Intelligence and\\
~~~~Data Processing Symposium (IDAP)\},\\
~~title=\{Development of Deep Learning Optimizers: Approaches,\\
~~~~Concepts, and Update Rules\},\\
~~year=\{2026\},\\
~~pages=\{1-19\},\\
~~doi=\{10.1109/IDAP71869.2026.11728119\}\}
}
\end{flushleft}
\vspace{0.1cm}
\hrule

\end{document}